\documentclass[10pt,twocolumn,letterpaper]{article}
\PassOptionsToPackage{table,dvipsnames}{xcolor}
\usepackage[pagenumbers]{iccv}
\usepackage{cuted}
\usepackage{caption}
\usepackage{amsmath,amssymb}
\usepackage{booktabs,multirow}
\usepackage{xcolor,colortbl}
\usepackage{graphicx}
\usepackage{enumitem}
\usepackage{xspace}
\usepackage{tikz}
\usepackage[skins]{tcolorbox}

\newcommand{\ours}{ArtKit\xspace}
\newcommand{\hart}{HArt\xspace}
\newcommand{\harto}{HArt-O\xspace}
\newcommand{\harth}{HArt-H\xspace}
\DeclareMathOperator*{\argmin}{arg\,min}

\definecolor{ourscolor}{RGB}{220,230,240}

\definecolor{cGen}{HTML}{1F7A74}
\definecolor{cObs}{HTML}{6F5BA8}
\definecolor{cSeg}{HTML}{C9841C}
\definecolor{cIdea}{HTML}{4A6B8A}
\newtcolorbox{ideabox}[1]{enhanced,colback=#1!14!white,colframe=#1!80!black,boxrule=0.5pt,arc=2.5pt,left=4pt,right=4pt,top=2.5pt,bottom=2.5pt,boxsep=0pt,before skip=3pt plus 1pt,after skip=3pt plus 1pt,fontupper=\footnotesize}
\newcommand{\gapbox}[3]{\begin{ideabox}{#1}\textbf{#2} #3\end{ideabox}}
\DeclareRobustCommand{\mk}[2]{\tikz[baseline=(m.base)]{\node[circle,draw=#2!80!black,fill=#2!20!white,line width=0.45pt,inner sep=0pt,minimum size=1.3em,font=\sffamily\bfseries\scriptsize,text=#2!50!black](m){#1};}}
\DeclareRobustCommand{\mG}{\mk{G}{cGen}\,}
\DeclareRobustCommand{\mO}{\mk{O}{cObs}\,}
\DeclareRobustCommand{\mS}{\mk{S}{cSeg}\,}
\newcommand{\mGt}{{\color{cGen}\textbf{[G]}}\,}
\newcommand{\mOt}{{\color{cObs}\textbf{[O]}}\,}
\newcommand{\mSt}{{\color{cSeg}\textbf{[S]}}\,}

\providecommand{\akSE}{\mathrm{SE}(3)}
\providecommand{\akSO}{\mathrm{SO}(3)}
\providecommand{\akExp}{\operatorname{Exp}}

\definecolor{linkblue}{rgb}{0.21,0.49,0.74}
\usepackage[pagebackref,breaklinks,colorlinks,allcolors=linkblue]{hyperref}

\title{ArtKit: Motion-Driven Re-Segmentation and Link Re-Generation for Articulated Objects from RGB-D Video}

\author{Sylvia Yuan$^{1,*}$ \qquad Dan Wang$^{2,*}$ \qquad Vivian Yu$^{1}$ \qquad Ravi Ramamoorthi$^{1}$ \qquad Xinrui Cui$^{3,\dagger}$\\[4pt]
$^{1}$University of California San Diego \qquad $^{2}$University of Arizona \qquad $^{3}$University of North Texas\\[2pt]
{\small $^{*}$Equal contribution \qquad $^{\dagger}$Corresponding author}}

\begin{document}
\maketitle
\begin{strip}
\centering
\includegraphics[width=\textwidth]{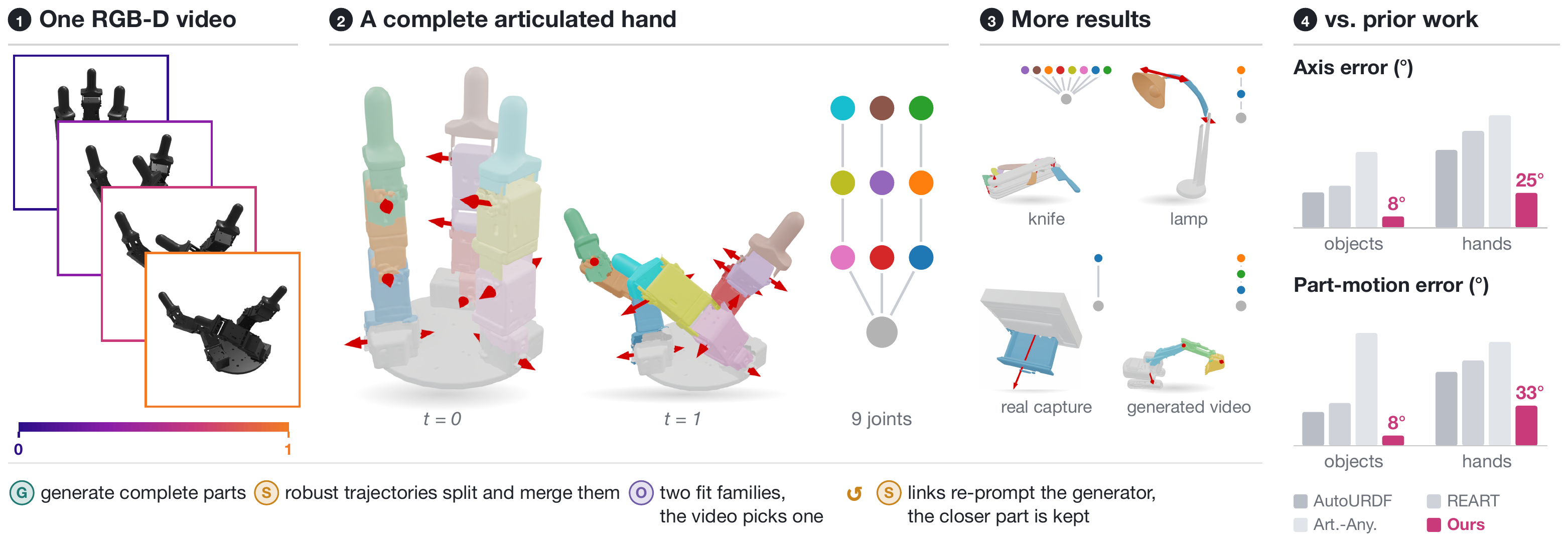}
\captionsetup{hypcap=false}
\captionof{figure}{\textbf{From one RGB-D video to a complete articulated object.} \ours takes an RGB-D video of an object being moved, here the nine-joint D'Claw gripper (1), and returns complete parts, typed joints and the articulation tree (2). The same pipeline handles household objects, real recordings and generated videos (3), and lowers the axis and part motion errors of prior work on \harto and \harth (4).}
\captionsetup{hypcap=true}
\label{fig:teaser}
\end{strip}

\begin{abstract}
An RGB-D camera that films an object being moved, often from one viewpoint, sees one side of it and how its parts move. We present \ours, which takes such a video as input to create an articulated model with complete parts, a kinematic tree, and typed joints. Our central idea is generating the hidden geometry and verifying it against what the camera sees. A generative part prior proposes complete, part-decomposed geometry based on the first frame. Point tracks, grouped into rigid trajectories, split and merge the generated parts into links. We then fit revolute and prismatic joints, with a visibility-aware energy ensuring that the articulated model explains the depth observed across the video. A rule checks against the video to choose between synchronous group fits and per-frame chain fits, so that one pipeline serves a diverse range of articulated objects, such as household items and robotic hands. Finally, we regenerate each link of the chosen fit as one part and keep it wherever it agrees better with the observations. We evaluate on Articulate-100 and on \hart, a new benchmark of 60 PartNet-Mobility objects and 8 robotic hands and grippers rendered as RGB-D videos with ground-truth parts and joints. \ours\ reduces the axis and part motion errors of the strongest baseline by more than half on the objects, as well as the axis error on the hands. \ours also runs on real recordings and generated videos.
\end{abstract}

\section{Introduction}

Articulated digital models extend an object's geometry to describe the mechanisms that move its parts. A cabinet requires doors of a given thickness on hinges with a given range of motion, and a robotic hand requires distinct links connected through a kinematic chain. Recovering such models from a short interaction video would simplify asset creation for robotics, simulation, and digital twins. However, a single camera is only able to observe non-occluded parts of the object's surface; moreover, the recorded interaction constrains only how parts involved in the interaction move. Therefore, reconstruction of complete geometry with rigid-link structure and articulation that agree with these partial, lossy observations is a challenging problem.

Video-based methods recover articulation from tracked motion, and generative methods supply complete shapes from static inputs. Joining them is not direct, because semantic parts need not coincide with rigid links, the pieces of an object that move as one. A generated lamp arm may fuse two links that are in fact separated by a joint, while its shade and bracket may be generated separately despite moving rigidly together. Fitting joints to a fixed decomposition cannot correct these mistakes, since one rigid mesh that corresponds to two moving parts cannot execute two independent motions irrespective of how the joint axis is fitted. Relying on similar trajectories alone handles ambiguous cases poorly, and can merge disconnected components into one erroneous link, such as fingers on a hand performing similar motion. Therefore, structural correction must combine evidence of relative motion with spatial association.

The links of an object form a tree, in which a lamp's head hangs on its arm and a fingertip on its finger. Hierarchy introduces a second difficulty. A child’s world-space trajectory contains motion inherited from every ancestor as well as its own joint. Consequently, grouping by total displacement is insufficient for determining parent-child relations. The relevant constraint is whether a candidate child’s motion relative to its parent is consistent with a single revolute or prismatic joint. Small excursions, self-occlusion, and synchronized motions weaken this evidence and can leave multiple structures compatible with the video. Among the links that touch, \ours picks the tree in which each child moves relative to its parent most like a single revolute or prismatic joint (Sec.~\ref{sec:reseg}).

We present \ours, a pipeline that treats generated parts as revisable geometric hypotheses. \ours\ uses generation and observation to correct each other, and we mark its three technical contributions throughout the paper and in Fig.~\ref{fig:framework}, \mG the generative part prior (Sec.~\ref{sec:prior}), \mS the motion-driven re-segmentation (Sec.~\ref{sec:reseg}) and \mO the observation-driven articulation (Sec.~\ref{sec:artic}).
\gapbox{cIdea}{Key idea.}{Generate the unseen, verify with the seen. The generative prior proposes complete parts, the observations decide how they move, and the recovered motion rewrites the parts, until the posed model is able to explain every observed depth map.}

We take as input an RGB-D video, with known camera poses and typically from a fixed viewpoint, of an object being moved. The part generator uses only the first frame, and the rest of the video is used to determine how the generated parts move and how they are cut into links. We assume the number of links $K$ to be known, while the kinematic tree, the joint types, the axes and the motion are inferred.

Our contributions are:
\begin{itemize}[nosep,leftmargin=*]
\item \mG\textbf{A generative part prior for articulation} (Sec.~\ref{sec:prior}). Complete, part-decomposed geometry generated from the first frame and matched to its depth, so that every part has a back, an inside and a plausible boundary before any motion is fitted.
\item \mS\textbf{Motion-driven re-segmentation} (Sec.~\ref{sec:reseg}). Point tracks are grouped into robust rigid trajectories, and tracks that no rigid body explains are left out. The trajectories split and merge the generated parts into links and give the kinematic tree. Once the articulation is selected, the links are used to re-prompt the part generator, and each link keeps either the regenerated or the original geometry, whichever better agrees with its observed points.
\item \mO\textbf{Observation-driven articulation} (Sec.~\ref{sec:artic}). We fit a chain of typed joints to the depth of every frame with a visibility-aware Chamfer energy and track and smoothness terms, recovering up to 20 joints per object. Synchronous group fits and per-frame chain fits are pooled, and a rule that uses only the video selects one, so objects and hands share one pipeline.
\item \textbf{\hart} (Sec.~\ref{sec:setup}). A new benchmark of 60 household objects and 8 robotic hands and grippers rendered as RGB-D videos with ground-truth parts and joints, scored by one protocol in which predicted and true parts are paired once by Chamfer distance, and unpaired joints and failed objects receive the maximum error.
\end{itemize}

Articulated datasets with complete part meshes are few, especially with joint chains. Thus, our quantitative evaluation is on rendered videos of Articulate-100~\cite{artmesh} and of \hart. \ours lowers the axis and part motion errors of the strongest baselines by a wide margin on household objects and on robotic hands, and also runs on real recordings and generated videos. Sec.~\ref{sec:related} reviews related work, Sec.~\ref{sec:method} describes \ours, and Sec.~\ref{sec:exp} evaluates it.

\section{Related Work}
\label{sec:related}

\textbf{Articulated objects from observations.} From two-state multi-view captures, PARIS~\cite{paris}, ArtGS~\cite{artgs}, GaussianArt~\cite{gaussianart} and ArtMesh~\cite{artmesh} fit part-aware radiance fields, Gaussians or meshes to both states and recover the rigid motion between them. DTA~\cite{digitaltwinart} registers two RGB-D scans of the two states, REART~\cite{reart} fits point cloud videos, Chao~\etal~\cite{gsart} segment a point cloud sequence with dynamic 3D Gaussians, and VideoArtGS~\cite{videoartgs} takes a monocular video whose camera moves around the object. Ditto~\cite{ditto} and OPD~\cite{opd,opdmulti} predict joints and part segments from the observed input, and A3VLM~\cite{a3vlm} predicts part boxes and axes from one image. The reconstruction methods recover only what the cameras saw, so with one fixed camera the unseen side of every part is missing. \ours keeps the observation-driven fit, \mO, applies it to complete generated parts, \mG, and lets the observed motion decide where those parts are cut and how the links form a tree, \mS, which recovers chains of up to 20 joints from one view.

\textbf{Generated articulated objects.} NAP~\cite{nap}, CAGE~\cite{cage} and SINGAPO~\cite{singapo} sample part graphs and joints from a learned distribution, URDFormer~\cite{urdformer} predicts the URDF structure from an image and places template parts, Real2Code~\cite{real2code} and Articulate-Anything~\cite{articulate-anything} write the articulation as code with language models, Articulate-AnyMesh~\cite{articulate-anymesh} articulates a given mesh with vision-language prompts, and DreamArt~\cite{dreamart} and Kinematify~\cite{kinematify} build articulated objects from images with video diffusion or geometric search. Their joints come from learned or language priors and static geometry, rather than from motion fitted to an observation. Image-to-3D models~\cite{hunyuan3d21,trellis} produce complete meshes from one picture, and 3D part models~\cite{p3sam,xpart,partfield} segment or regenerate them as parts. \ours uses Hunyuan3D-2.1~\cite{hunyuan3d21} and X-Part~\cite{xpart} as a prior whose geometry and decomposition the observations place, correct and re-cut, and takes every joint from the observations.

\textbf{Robotic hands.} Most articulation benchmarks contain furniture-like objects with few joints, and recent work reaches about twenty parts on household objects~\cite{gaussianart}. Hands and grippers have many joints in a deep tree, with small, similar, self-occluding parts, and Kinematify~\cite{kinematify} synthesizes such high-DoF objects from images or text. \harth, built from dex-urdf~\cite{dexurdf}, benchmarks such hands.

\section{Method}
\label{sec:method}

\begin{figure*}[t]
\centering
\includegraphics[width=\textwidth]{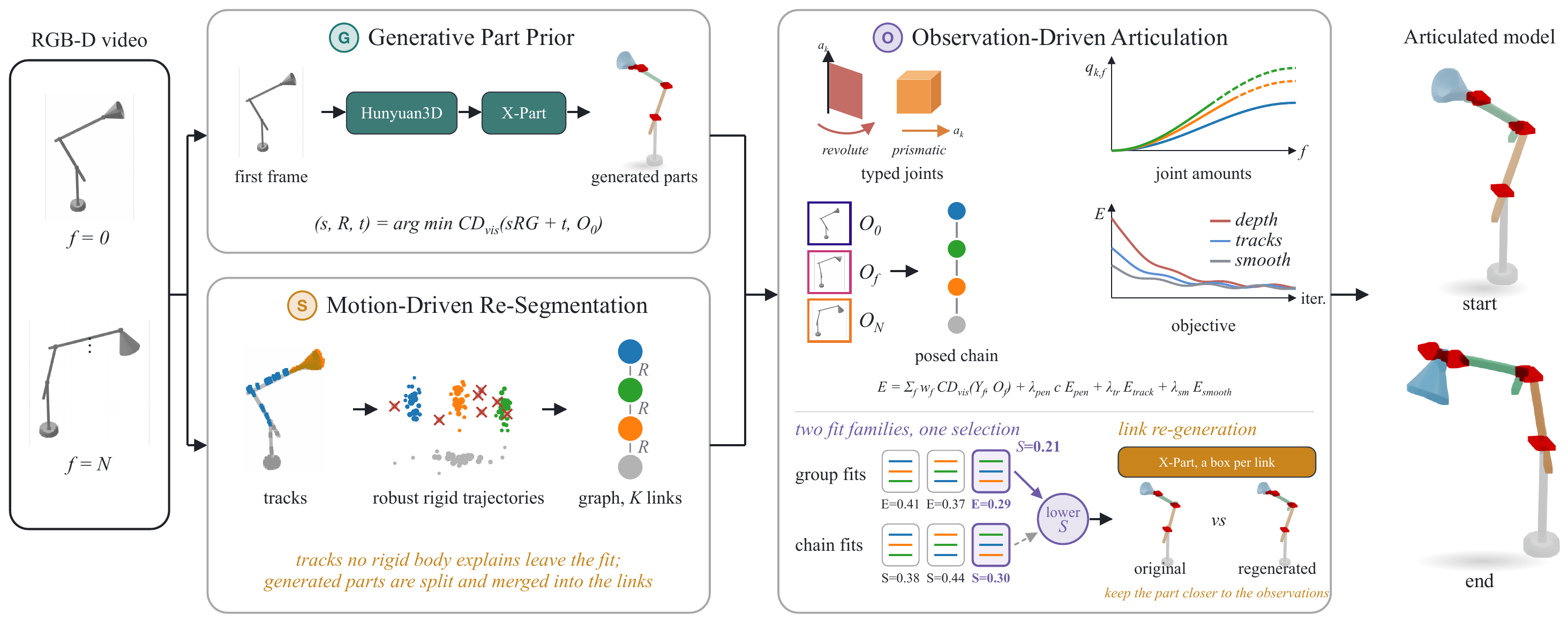}
\caption{\textbf{Overview.} \mGt \emph{Generative part prior} (Sec.~\ref{sec:prior}). Hunyuan3D and X-Part propose complete parts from the first frame, and a visibility-aware Chamfer fit places them into $O_0$. \mSt \emph{Motion-driven re-segmentation} (Sec.~\ref{sec:reseg}). Point tracks are grouped into robust rigid trajectories, with tracks that no rigid body explains left out, the generated parts are split and merged into links, and the links give the articulation graph. \mOt \emph{Observation-driven articulation} (Sec.~\ref{sec:artic}). Typed joints are fitted to the depth of every frame with track and smoothness terms by a pool of synchronous group fits and per-frame chain fits, each giving per-frame joint amounts $q_{k,f}$. The video alone is used to select one fit. Each selected link then re-prompts X-Part with its own box, and the regenerated part is kept where it fits the link's points better.}
\label{fig:framework}
\end{figure*}

\textbf{Problem formulation.} The input is an RGB-D video $\{(I_f,D_f,M_f)\}_{f=0}^{N}$ of an object moving, with an object mask $M_f$ per frame and known camera poses $C_f\in\akSE$ that map world to camera coordinates. The first frame shows the object at rest; this is the only frame the generative prior sees. Every pixel $u$ is lifted to the world by its depth,
\begin{equation}
 x(u,f)=C_f^{-1}\bigl(D_f(u)\,A^{-1}\widetilde u\bigr),\qquad \widetilde u=(u_x,u_y,1)^\top,
 \label{ak:eq:lift}
\end{equation}
with intrinsics $A$, so a moving camera and a fixed one present the same world-frame observations $O_f=\{x(u,f):M_f(u)=1\}$ (our benchmarks use a fixed camera). A link is a rigid piece of the object, and a joint connects a link to its parent link. We assume the number of links $K$, including the static base, to be known; the tree, the joint types, axes and motions are inferred. The output is a complete mesh $P_k$ per link, a tree with parent map $\pi$ rooted at the stationary base $P_0$, a joint per moving link with type $\tau_k\in\{\mathrm{R},\mathrm{P}\}$ and screw axis $\xi_k$, and an amount $q_{k,f}$ per link and frame with $q_{k,0}=0$. A link's world pose follows the tree,
\begin{equation}
 T_{k,f}=T_{\pi(k),f}\,\akExp\bigl(\widehat\xi_k\,q_{k,f}\bigr),\qquad T_{0,f}=I,
 \label{ak:eq:fk}
\end{equation}
where $\xi_k=(a_k,-a_k\times o_k)$ for a revolute joint about the line through $o_k$ along the unit direction $a_k$, with $q$ an angle, and $\xi_k=(0,a_k)$ for a prismatic joint, with $q$ a distance. We write $Y_f(q,\xi)=\bigcup_k T_{k,f}P_k$ for the model posed at frame $f$. All distances are measured relative to the length $L$, the diagonal of the bounding box of $O_0$, which is fixed once per object, and $\varepsilon=0.02L$ is the surface tolerance.

\textbf{Visibility-aware energy.} Every hypothesis is scored against a frame by a visibility-aware Chamfer distance. For model points $X$ and observed points $O$,
\begin{equation}
 \begin{split}
 \mathrm{CD}_{\mathrm{vis}}(X,O)={}&\frac{1}{|O|}\sum_{o\in O}\min\bigl(d(o,X),c\bigr)\\
 &+\frac{1}{|V(X)|}\sum_{x\in V(X)}\min\bigl(d(x,O),c\bigr),
 \end{split}
 \label{ak:eq:cdvis}
\end{equation}
where $d(\cdot,\cdot)$ is the distance to the nearest point and $c=0.1L$ caps outliers. $V(X)$ keeps the points of $X$ that the camera of frame $f$ sees, \ie those not hidden behind another point of $X$ along their camera ray. The first term asks that every observed point be covered by the model. The second asks only the visible model to lie on the observation, so the back of a part, which the camera cannot see, is not penalized.

\textbf{Overview.} \ours runs one pipeline on every input (Fig.~\ref{fig:framework}). The prior \mG proposes complete parts from the first frame and places them into $O_0$ (Sec.~\ref{sec:prior}). The motion \mS groups point tracks into rigid trajectories, turns the generated parts into $K$ links and gives the tree (Sec.~\ref{sec:reseg}). The articulation \mO fits the joints to every frame (Sec.~\ref{sec:artic}). Grouping and fitting are not convex, so every input receives a pool of fits from two families. Group fits move all joints together along one progress through the clip, and chain fits give every joint its own amount in every frame. Chain fits are more flexible, but they cut their own links and have many more unknowns, so on joints that move together the tied group fit is more reliable. No threshold decides between them; the video does. A rule that uses the video alone selects one fit (Eq.~\eqref{ak:eq:select}). The links of the selected fit then re-prompt the part generator, and each regenerated link is kept where it is closer to the observations.

\subsection{\texorpdfstring{\mGt}{G} Generative Part Prior}
\label{sec:prior}

\textbf{Complete parts from one frame.} This step is \mG in Fig.~\ref{fig:framework}. A single depth map is a one-sided shell, and articulation needs information about the whole part, since the whole part is what moves and lands on the observed depth once it has been posed. Hunyuan3D-2.1~\cite{hunyuan3d21} generates a complete mesh of the object in $I_0$, and X-Part~\cite{xpart} cuts it into complete parts $\{G_j\}_{j=1}^{J}$, with part boxes proposed by P3-SAM~\cite{p3sam}. For the hands, whose fingers this tends to fuse, a second and finer decomposition is made with more P3-SAM prompts, and both enter the pool (Sec.~\ref{sec:artic}). The parts, which follow the object's appearance, are posed in the generator's own frame, and need not be cut along the links. A similarity transform places the generated points $G=\bigcup_j G_j$ into the world,
\begin{equation}
 (s^\star,R^\star,t^\star)=\argmin_{s>0,\,R\in\akSO,\,t}\ \mathrm{CD}_{\mathrm{vis}}\bigl(sRG+t,\;O_0\bigr),
 \label{ak:eq:alignment}
\end{equation}
minimized by ICP from 288 starts, the 24 axis-aligned orientations each turned 12 ways about the vertical, of which the eight best are refined. Each part is then moved rigidly on its own to fit $D_0$ more closely. The appendix gives the details.

\subsection{\texorpdfstring{\mSt}{S} Motion-Driven Re-Segmentation}
\label{sec:reseg}

\textbf{Rigid trajectories.} X-Part cuts the mesh where its shape changes, for example between a lamp's shade and its arm, while the links of the object are separated by its joints. So a finger may be generated as one part across two phalanges, or a lamp's arm cut in the middle. We use the motion in the video to decide which pieces move together (\mS in Fig.~\ref{fig:framework}, from the tracks to the links). Point tracks follow every object pixel of the first frame through the video, by chained optical flow~\cite{raft} for the group fits and by CoTracker3~\cite{cotracker3}, forward and backward, for the chain fits (Sec.~\ref{sec:artic}). A track $n$ lifted by Eq.~\eqref{ak:eq:lift} gives points $x_{n,f}$. Sequential RANSAC groups the tracks into $B$ rigid trajectories $H_{b,f}\in\akSE$, $b=1,\dots,B$, with $H_{b,0}=I$, and $\bar r_{nb}$ is the mean squared distance of track $n$ from $H_{b,f}\,x_{n,0}$ over its frames, divided by $L^2$.

\textbf{Robust rigid trajectories.} Given the $B$ trajectories, each track must be assigned to one of them. A track that slips across a boundary or drifts on a textureless surface follows no rigid body, and a hard assignment lets such a track pull the trajectory it is assigned to. We therefore weigh tracks by soft memberships in a mixture of the $B$ trajectories and an outlier component $\varnothing$,
\begin{equation}
 \gamma_{nb}\propto\frac{1-\eta_0}{B}\,e^{-\bar r_{nb}/2\sigma^2},\qquad \gamma_{n\varnothing}\propto\eta_0\,e^{-\rho^2/2\sigma^2},
 \label{ak:eq:membership}
\end{equation}
with inlier radius $\rho$ in units of $L$, $\sigma=\rho/2$ and outlier weight $\eta_0=0.05$. Each track is thus shared among the trajectories by how well each one carries it, and the outlier component takes most of the weight of a track that no trajectory carries closely. Robust track cleaning leaves out every track with $\gamma_{n\varnothing}>\tfrac12$. Robust trajectories refit each trajectory from all tracks weighted by their memberships, alternating with Eq.~\eqref{ak:eq:membership} for three rounds, so a slipping track barely moves it. Track cleaning and robust trajectories each add fits to the pool (Sec.~\ref{sec:artic}).

\textbf{Split, merge and the tree.} Each generated point $g$ takes the trajectory most common among the tracks that start nearest to it, $\beta(g)=\operatorname{mode}\{b_n\}$, where $b_n$ is the trajectory carrying track $n$. A generated part is cut along the boundary between its points of different $\beta$, parts whose points share one $\beta$ are merged, and separate pieces with one $\beta$, such as two fingers, stay separate links. During the joint search (Sec.~\ref{sec:artic}), a region whose depth or tracks contradict its link's fitted motion is split off if the energy $E$ of Eq.~\eqref{ak:eq:energy} drops to $E'\le(1-\delta_s)E$, and two touching links are merged if it rises only to $E'\le(1+\delta_m)E$. Hanging link $k$ on a touching link $p$ costs $\kappa(p,k)$, the residual of the best single revolute or prismatic joint at their contact that explains the relative motion $\Delta_{pk,f}=T_{p,f}^{-1}T_{k,f}$, since a child moves about one fixed axis relative to its true parent only. The tree is the minimum-cost arborescence rooted at the base, $\pi=\argmin_{\pi}\sum_k\kappa(\pi(k),k)$ (appendix).

\textbf{The links re-prompt the generator.} A link merged from several generated parts is a set of pieces with seams between them. Once the articulation has been selected (Sec.~\ref{sec:artic}), we map each link's points back into the generator's frame by the inverse of the placement of Eq.~\eqref{ak:eq:alignment}, and give their bounding box to X-Part together with the same complete mesh. Since X-Part draws one part per box, each link is returned as one whole part $P^{\mathrm{new}}_k$ (the arrow from \mS back to \mG in Fig.~\ref{fig:framework}). The regenerated part is a second hypothesis for the link, and we choose, based on the observations,
\begin{equation}
 P_k=\begin{cases}P^{\mathrm{new}}_k & \text{if }\ \mathrm{CD}\bigl(P^{\mathrm{new}}_k,X_k\bigr)<\mathrm{CD}\bigl(P^{\mathrm{old}}_k,X_k\bigr),\\ P^{\mathrm{old}}_k & \text{otherwise,}\end{cases}
 \label{ak:eq:linkpick}
\end{equation}
for every link $k$ of the selected fit, where $X_k$ are the link's fitted points, $P^{\mathrm{old}}_k$ is the link assembled from the original parts and $\mathrm{CD}$ is the symmetric Chamfer distance. The joints and poses are kept, and the rule is the same for every input. The motion decides the decomposition the generator draws, and the observations decide whether the redrawn part replaces the original.

\subsection{\texorpdfstring{\mOt}{O} Observation-Driven Articulation}
\label{sec:artic}

\textbf{A posed model must explain each frame.} This step is \mO in Fig.~\ref{fig:framework}. A hypothesis about the joints is judged by the depth it should produce at the frames of the video. The articulation minimizes
\begin{equation}
 \begin{split}
 E(q,\xi)={}&\sum_{f}w_f\,\mathrm{CD}_{\mathrm{vis}}\bigl(Y_f(q,\xi),O_f\bigr)+\lambda_{\mathrm{pen}}\,c\,E_{\mathrm{pen}}\\
 &+\lambda_{\mathrm{tr}}E_{\mathrm{track}}+\lambda_{\mathrm{sm}}E_{\mathrm{smooth}},
 \end{split}
 \label{ak:eq:energy}
\end{equation}
with frame weights $w_f$, a track term that charges every track for its distance from where its link carries it, a smoothness term on the second differences of the amounts $q_{k,f}$ over the frames, and a penetration term, the fraction of moving points deep inside another link, scaled by the cap $c$ of Eq.~\eqref{ak:eq:cdvis} (appendix). Each joint is initialized from a set of candidates: hinges along the edges and symmetry axes of the link's bounding box, slides along the box axes, and the screw that best explains the link's tracked motion. The candidate with the lowest $E$ is then refined by ICP, with its amount free.

\textbf{Two families of fits.} No single kind of fit works on every object. When the joints move together, as when a lamp is raised, one progress shared by all joints, scaled by a final amount per joint, explains the whole clip and leaves few unknowns. In contrast, deep chains of small, self-occluding links, such as fingers, are better determined frame by frame, where every joint's amount is free in every frame and is constrained by that frame's depth and by long point tracks. A synchronous group fit therefore ties every joint to one progress through the clip, $q_{k,f}=\theta_f\,q_k$ with $\theta_f=f/N$. This linear progress approximates an uneven pace, which the lower weight of the in-between frames in $E$ tolerates; joints moving at different paces are left to the chain fits. A per-frame chain fit segments the CoTracker3 tracks into rigid bodies by their motion, leaves every $q_{k,f}$ free, and refines all joints together by gradient descent on Eq.~\eqref{ak:eq:energy}. Each family is run several times, with perturbed track settings and seeds and with the robust trajectories of Sec.~\ref{sec:reseg}, on the part decompositions of Sec.~\ref{sec:prior} (appendix). The appendix lists the settings of every fit in the pool.

\textbf{Choosing among fits.} The fits differ in track settings, seeds and robust variants, and chain fits cut their own links, so the pool holds several candidate models. Group fits on the same part decomposition share their links, and can therefore be compared by $E$. In contrast, each chain fit segments the tracks on its own, so two chain fits, or a chain fit and a group fit, may split the object into different links. Since the two families also weigh the frames and terms of $E$ differently, their energies are not comparable. Such fits are compared by a simpler check against the video. Each fit's links are posed at every frame, and
\begin{equation}
 S=\frac{1}{N+1}\sum_{f}\Bigl(\frac{|Y_f\cap\Omega_f|}{|Y_f|}+\frac{|\{o\in O_f:\ d(o,Y_f)>\varepsilon\}|}{|O_f|}\Bigr)
 \label{ak:eq:check}
\end{equation}
counts the posed points that fall in $\Omega_f$, the space frame $f$ shows to be empty (in front of the observed depth or outside $M_f$), and the observed surface left unexplained. A fit that puts no posed point where the camera sees empty space and explains every observed point has $S=0$. With $\mathcal M^{\mathrm{grp}}_g$ the group fits on part decomposition $g$ and $\mathcal M^{\mathrm{ch}}$ the chain fits, the group fits on each decomposition propose their fit with the lowest $E$, the chain fits propose their fit with the lowest $S$, and these finalists are then compared on $S$,
\begin{equation}
 \begin{split}
 &\hat m_g=\argmin_{m\in\mathcal M^{\mathrm{grp}}_g}E(m),\qquad \hat m^{\mathrm{ch}}=\argmin_{m\in\mathcal M^{\mathrm{ch}}}S(m),\\
 &m^\star=\argmin_{m\in\{\hat m_g\}_g\cup\{\hat m^{\mathrm{ch}}\}}S(m).
 \end{split}
 \label{ak:eq:select}
\end{equation}
Thus, $E$ ranks group fits that share their links, whereas $S$ ranks all other fits. The choice uses the video alone. The links of $m^\star$ then re-prompt the generator (Eq.~\eqref{ak:eq:linkpick}). The output is a set of complete parts, the tree, the joints and the per-frame poses, which can be posed beyond the observed motion and exported as a URDF.

\section{Experiments}
\label{sec:exp}

\subsection{Setup}
\label{sec:setup}

\textbf{Articulate-100.} Articulate-100~\cite{artmesh} holds 100 PartNet-Mobility~\cite{sapien} objects across 16 categories with from two to over six parts, with every movable part attached directly to the base. We render and score each object as for \hart below.

\begin{figure}[t]
\centering
\includegraphics[width=0.8\linewidth]{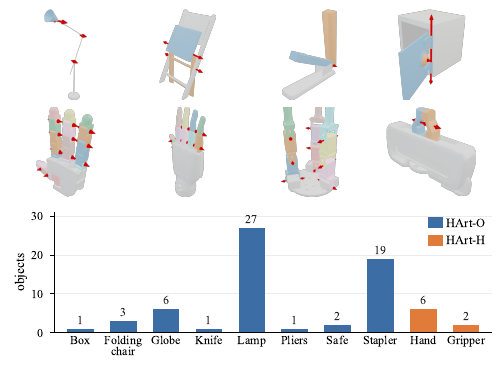}
\caption{\textbf{\hart.} Top: four \harto objects and four \harth hands and grippers at rest, parts colored and joint axes in red. Bottom: objects per category in \harto, and the six hands and two grippers of \harth.}
\label{fig:dataset}
\end{figure}

\textbf{\hart.} Few articulated datasets come with complete part meshes, and in most PartNet-Mobility objects some part, such as a drawer's inside compartment or a door behind another, is occluded from a single camera, which can make it difficult for any method to recover from one view. We therefore introduce \hart, a benchmark of articulated objects observed by one fixed RGB-D camera, in two halves. \harto holds 60 PartNet-Mobility~\cite{sapien} objects from eight categories (appendix), with one to six joints per object, most of them in chains, and 159 joints in total. \harth holds 8 dexterous hands and grippers from dex-urdf~\cite{dexurdf,bunnyvisionpro}, the Allegro~\cite{allegrohand}, Schunk SVH~\cite{schunkhand}, Ability~\cite{abilityhand}, LEAP~\cite{leaphand}, Inspire~\cite{inspirehand} and Barrett~\cite{barretthand} hands, the D'Claw~\cite{robel} and the Panda gripper~\cite{franka}, with 2 to 20 joints. Every object is rendered by a fixed camera as a 30-frame RGB-D video with object masks, starting at rest, in which every joint moves a part of its range. Ground truth is the posed part mesh of every state and the joint axes, types and motions. Fig.~\ref{fig:dataset} shows examples. We will release \hart.

\textbf{Protocol.} Predicted and ground-truth parts are paired once, by a Hungarian assignment on their Chamfer distance averaged over the start and end state; this pairing scores both geometry and joints. Each ground-truth joint is scored through its paired part by the axis angle, the distance between the axis lines in units of 0.1\,m, and the part motion error at the end state. A predicted joint is read as a hinge when its part turns by $5^\circ$ or more and as a slide otherwise. A wrong predicted type receives the motion penalty, and a predicted slide on a true hinge receives the axis position penalty. Geometry is the Chamfer distance of the static part (CD-s) and the movable parts (CD-m) in mm, averaged over the start and end states, with 1000\,mm for an unpaired part. An unpaired joint, a paired part that stays static, or a failed object receives the maximum error ($90^\circ$, 10, and $90^\circ$ or 1\,m)~\cite{video2articulation}. AutoURDF and REART receive the ground-truth part count.

\textbf{Baselines.} AutoURDF~\cite{autourdf} registers clusters of a point cloud sequence into a URDF, REART~\cite{reart} fits a rearticulable model to a 4D point cloud, and Articulate-Anything~\cite{articulate-anything} retrieves an object mesh from PartNet-Mobility and writes its articulation as URDF code from the video with a vision-language model. AutoURDF and REART receive point cloud sequences of the posed ground-truth meshes and the part count (appendix), and Articulate-Anything receives the video. On Articulate-100 we also report VideoArtGS~\cite{videoartgs}, which uses a moving camera and attaches every part directly to the base, and therefore cannot represent the joint chains of \hart.

\begin{figure*}[t]
\centering
\includegraphics[width=\textwidth]{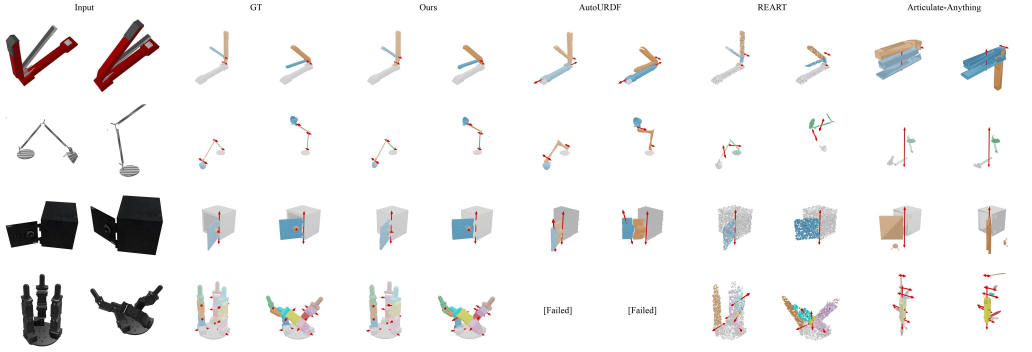}
\caption{\textbf{Articulation on \harto and \harth.} Each method is shown at the start and end state of the input video, rendered in the same frame as the ground truth, with parts colored by their matched ground-truth part and the predicted joint axes drawn as red arrows; [Failed] marks a method that returned no model. \ours\ recovers the hinge chains of the stapler, lamp and safe and the D'Claw's fingers with complete parts, whereas AutoURDF and REART work on point clouds of the ground-truth surface, and Articulate-Anything returns a retrieved object whose joints follow the retrieval (for the hand, from a library of hands without the tested one).}
\label{fig:qual}
\end{figure*}

\subsection{Comparison}

\begin{table}[t]
\centering
\caption{\textbf{Comparison on Articulate-100} (100 objects, every movable part attached to the base). Articulation: axis angle ($^\circ$), axis position (0.1\,m), part motion ($^\circ$ or m), each averaged over an object's movable parts and then over objects. Geometry: Chamfer distance of the static and movable parts (mm). Unmatched joints and failed objects receive the penalty. $^\dagger$VideoArtGS receives 172 frames per object from a moving camera with ground-truth depth and camera poses, 100 static views of the object at rest followed by 72 frames in which all joints move together, and joint types predicted by GPT-4o given the ground-truth part count, and is scored under the same protocol. Best in \textbf{bold}, second \underline{underlined}. Gain is the improvement over the best baseline in each column.}
\label{tab:art100}
\setlength{\tabcolsep}{3pt}
\resizebox{\linewidth}{!}{
\begin{tabular}{lcccccc}
\toprule
\multirow{2}{*}{Method} & \multicolumn{3}{c}{Articulation} & \multicolumn{2}{c}{Geometry} & \multirow{2}{*}{Fail $\downarrow$} \\
\cmidrule(lr){2-4} \cmidrule(lr){5-6}
 & Axis Ang $\downarrow$ & Axis Pos $\downarrow$ & Part Motion $\downarrow$ & CD-s $\downarrow$ & CD-m $\downarrow$ & \\
\midrule
AutoURDF~\cite{autourdf} & 33.70 & \underline{0.73} & \underline{6.26} & 185.3 & 203.6 & 1/100 \\
REART~\cite{reart} & 21.57 & 1.36 & 6.47 & \underline{44.0} & 170.2 & 0/100 \\
Articulate-Anything~\cite{articulate-anything} & 27.21 & 4.11 & 45.91 & 145.2 & 469.3 & 0/100 \\
VideoArtGS$^\dagger$~\cite{videoartgs} & \underline{16.77} & 1.41 & 10.31 & 57.5 & \underline{133.8} & 1/100 \\
\rowcolor{ourscolor} \ours & \textbf{7.92} & \textbf{0.32} & \textbf{1.99} & \textbf{42.9} & \textbf{74.3} & 0/100 \\
\midrule
\rowcolor{gray!12} Gain & +8.85 & +0.41 & +4.27 & +1.1 & +59.5 & \\
\bottomrule
\end{tabular}}
\end{table}

\textbf{Objects with one level of joints.} Table~\ref{tab:art100} compares all methods on Articulate-100, where every moving part is attached directly to the base. \ours has the lowest error in every column, with an axis error of $7.92^\circ$ against $16.77^\circ$ for VideoArtGS.

\begin{table}[t]
\centering
\caption{\textbf{Comparison on \harto} (60 objects, 159 joints). Articulation: axis angle ($^\circ$), axis position (0.1\,m), part motion ($^\circ$ or m). Geometry: Chamfer distance of the static part and the movable parts (mm). Unmatched joints and failed objects receive the penalty. Best in \textbf{bold}, second \underline{underlined}. Gain is the improvement over the best baseline in each column.}
\label{tab:main}
\setlength{\tabcolsep}{3pt}
\resizebox{\linewidth}{!}{
\begin{tabular}{lcccccc}
\toprule
\multirow{2}{*}{Method} & \multicolumn{3}{c}{Articulation} & \multicolumn{2}{c}{Geometry} & \multirow{2}{*}{Fail $\downarrow$} \\
\cmidrule(lr){2-4} \cmidrule(lr){5-6}
 & Axis Ang $\downarrow$ & Axis Pos $\downarrow$ & Part Motion $\downarrow$ & CD-s $\downarrow$ & CD-m $\downarrow$ & \\
\midrule
AutoURDF~\cite{autourdf} & \underline{25.02} & \underline{3.36} & \underline{28.10} & 186.8 & \underline{227.2} & 7/60 \\
REART~\cite{reart} & 29.69 & 3.41 & 35.59 & \underline{106.6} & 241.2 & 0/60 \\
Articulate-Anything~\cite{articulate-anything} & 53.37 & 8.32 & 93.72 & 557.4 & 613.6 & 2/60 \\
\rowcolor{ourscolor} \ours & \textbf{8.13} & \textbf{0.60} & \textbf{8.47} & \textbf{26.1} & \textbf{45.6} & 0/60 \\
\midrule
\rowcolor{gray!12} Gain & +16.89 & +2.76 & +19.63 & +80.5 & +181.6 & \\
\bottomrule
\end{tabular}}
\end{table}

\textbf{Hierarchical objects.} Most \harto and \harth objects carry chains of joints, so a part's motion mixes that of its own joint with the motion of every joint above it. On \harto (Table~\ref{tab:main}) \ours has the lowest error on every articulation metric by a wide margin, and the lowest Chamfer distances, with a movable-part error of 45.6\,mm against 227.2\,mm for the best baseline. Complete geometry matters most when a lid or a door swings into space that a reconstructed shell leaves empty, since only a part with a back places the hinge at its real edge. Articulate-Anything retrieves plausible objects of the right category, whose parts and axes rarely match the observed ones. Fig.~\ref{fig:qual} shows representative objects, and Fig.~\ref{fig:supp_qual} in the appendix five more.

\begin{figure*}[t]
\centering
\includegraphics[width=0.75\textwidth]{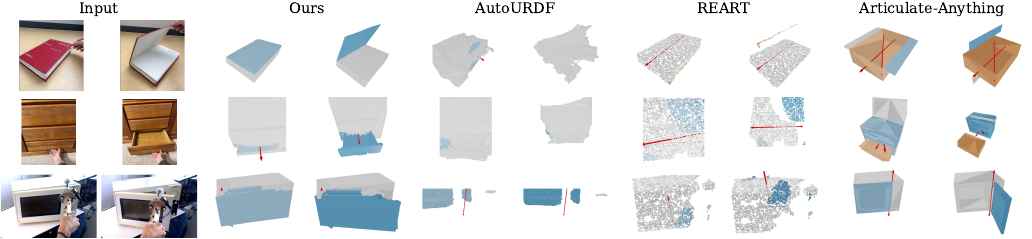}
\caption{\textbf{Real captures} at the start and end state, with joint axes in red. A book and a drawer from handheld iTACO captures and a microwave from RBO, seen from the input camera. \ours moves the cover, the drawer and the microwave front about their joints, whereas AutoURDF and REART lose most of the moving parts, and Articulate-Anything retrieves template objects whose shapes differ from the observed ones.}
\label{fig:real}
\end{figure*}

\begin{figure*}[t]
\centering
\includegraphics[width=0.70\linewidth]{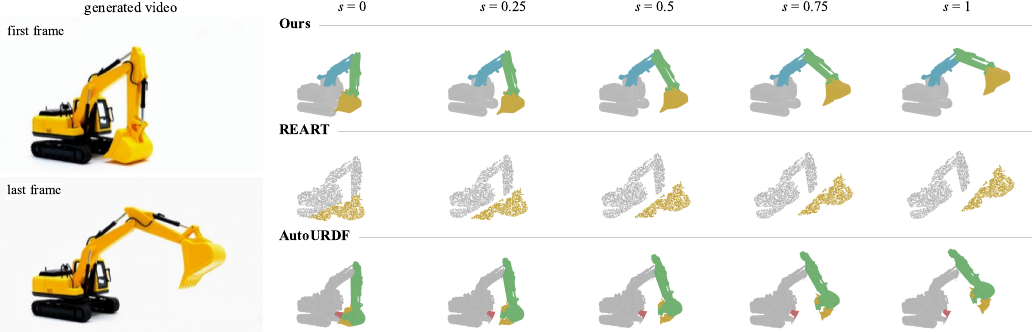}
\caption{\textbf{Articulated objects from generated videos.} First and last frame of a Wan2.1 clip, then \ours, REART and AutoURDF at five states $s$ of the motion.}
\label{fig:t2v}
\end{figure*}

\begin{table}[t]
\centering
\caption{\textbf{Robotic hands, \harth} (8 hands and grippers). The protocol of Table~\ref{tab:main}, 93 joints, unmatched joints penalized. Best in \textbf{bold}, second \underline{underlined}. Gain is the improvement over the best baseline in each column.}
\label{tab:hands}
\setlength{\tabcolsep}{3.5pt}
\resizebox{\linewidth}{!}{
\begin{tabular}{lcccccc}
\toprule
\multirow{2}{*}{Method} & \multicolumn{3}{c}{Articulation} & \multicolumn{2}{c}{Geometry} & \multirow{2}{*}{Fail $\downarrow$} \\
\cmidrule(lr){2-4} \cmidrule(lr){5-6}
 & Axis Ang $\downarrow$ & Axis Pos $\downarrow$ & Part Motion $\downarrow$ & CD-s $\downarrow$ & CD-m $\downarrow$ &  \\
\midrule
AutoURDF~\cite{autourdf} & \underline{54.73} & \underline{4.35} & \underline{61.41} & 463.1 & \underline{444.1} & 3/8 \\
REART~\cite{reart} & 68.09 & 6.62 & 70.91 & \underline{37.8} & 473.6 & 0/8 \\
Articulate-Anything~\cite{articulate-anything} & 79.18 & 9.07 & 86.35 & 733.1 & 786.2 & 2/8 \\
\rowcolor{ourscolor} \ours & \textbf{24.67} & \textbf{1.01} & \textbf{33.27} & \textbf{32.8} & \textbf{62.3} & 0/8 \\
\midrule
\rowcolor{gray!12} Gain & +30.06 & +3.34 & +28.14 & +5.0 & +381.8 & \\
\bottomrule
\end{tabular}}
\end{table}

\textbf{Robotic hands.} Our selection rule (Eq.~\eqref{ak:eq:select}) picks a per-frame chain fit for 6 of the 8 hands and for 16 of the 60 \harto objects. On \harth (Table~\ref{tab:hands}) \ours finds every scored joint and has the lowest error in every column, with less than half the axis error of the best baseline. Table~\ref{tab:supp_hands} in the appendix lists each hand; ranking by $S$ or $E$ alone raises the \harto axis error to $10.04^\circ$ or $10.68^\circ$.

\subsection{Ablations}

\begin{table}[t]
\centering
\caption{\textbf{Ablations} on \hart (the 60 objects of \harto and the 8 hands and grippers of \harth, each averaged over its joints and parts and then over the 68 objects; axis position over the 67 with a revolute joint), scored by the protocol of Table~\ref{tab:main}: axis angle ($^\circ$), axis position (0.1\,m), part motion ($^\circ$ or m), and Chamfer distance of the static and movable parts (mm). Each row removes one component; group fits only and chain fits only keep one family of the pool and select within it as in Eq.~\eqref{ak:eq:select}. Without robust trajectories the eight robust fits are replaced by eight fits with new seeds, so every row selects from the same number of fits. The full row is the mean of the \ours rows of Tables~\ref{tab:main} and~\ref{tab:hands} over the 68 objects. Best in \textbf{bold}, second \underline{underlined}.}
\label{tab:ablation}
\setlength{\tabcolsep}{3pt}
\resizebox{\linewidth}{!}{
\begin{tabular}{lccccc}
\toprule
 & Axis Ang $\downarrow$ & Axis Pos $\downarrow$ & Part Motion $\downarrow$ & CD-s $\downarrow$ & CD-m $\downarrow$ \\
\midrule
\mG single generated part & 23.47 & 2.34 & 29.52 & 70.9 & 163.5 \\
\mS without re-segmentation & 12.61 & 0.81 & 14.34 & \underline{28.8} & \textbf{45.6} \\
\mS without robust trajectories & \underline{11.80} & \underline{0.66} & \underline{12.41} & 35.3 & \underline{47.4} \\
\mO without video & 34.29 & 2.05 & 64.15 & 48.8 & 75.8 \\
\mO group fits only & 12.12 & 0.94 & 14.58 & 29.5 & 55.8 \\
\mO chain fits only & 21.83 & 1.49 & 29.54 & 58.3 & 103.2 \\
\midrule
\rowcolor{ourscolor} \ours (full) & \textbf{10.08} & \textbf{0.65} & \textbf{11.39} & \textbf{26.9} & 47.5 \\
\bottomrule
\end{tabular}}
\end{table}

In Table~\ref{tab:ablation}, we remove one component at a time on \harto and \harth together. When the generated mesh is kept as a single part, \mG, segmentation is left to the motion alone, and the axis error rises from $10.08^\circ$ to $23.47^\circ$. Without re-segmentation, \mS, the fused or over-cut parts of pliers, lamps and folding chairs articulate incorrectly, which raises the motion error to $14.34^\circ$. When the eight robust fits are replaced by eight fits with new seeds, the axis error rises to $11.80^\circ$. Link re-generation leaves the joints unchanged, and replaces 57\% of the moving links on \harto with parts made of fewer pieces, 2.84 against 3.33 components per link at nearly the same Chamfer distance (Table~\ref{tab:supp_regen_quality} and Fig.~\ref{fig:ablation_qual} in the appendix). Without the video, \mO, the fit only sees the start and end depth, which leaves the axis of a small rotation poorly determined. Finally, keeping only the group fits or only the chain fits raises the axis error to $12.12^\circ$ and $21.83^\circ$, respectively.

\subsection{Real Recordings and Generated Videos}

\textbf{Real recordings.} RBO~\cite{rbo} records people moving fourteen articulated objects in front of an RGB-D camera, fixed in the takes we use, with motion capture of every part, and iTACO~\cite{video2articulation} records people opening a book and pieces of furniture with a handheld phone that logs depth and camera poses. Since hands and arms cover the object in many frames, we generate the object from the first frame, from the first frame with the person and background removed by an image editing model~\cite{qwenimage} under one fixed prompt, or from the reversed clip. Fig.~\ref{fig:real} shows two iTACO captures and one RBO sequence. On the 17 RBO sequences with motion capture, \ours has the lowest axis angle, part motion and Chamfer errors (axis $32.45^\circ$ against $40.33^\circ$, CD-m 41.0 against 80.4\,mm), and AutoURDF the lowest axis position (Table~\ref{tab:supp_real} in the appendix; texture in Fig.~\ref{fig:texture}).

\noindent\textbf{Generated videos.} We generate a toy excavator unfolding its arm with the first-last-frame model of Wan2.1 (14B)~\cite{wan2025}, estimate depth and cameras with Depth Anything 3~\cite{depthanything3}, and mask the object with SAM 3~\cite{sam3}. REART and AutoURDF receive the depth as point clouds. In Fig.~\ref{fig:t2v} \ours recovers the boom, stick and bucket as one chain. REART keeps one moving part, and AutoURDF leaves the boom still.

\section{Conclusion}

\ours turns a short RGB-D video into a complete articulated object by utilizing generation and observation to correct each other. A generative prior supplies the proposed parts, the observations decide how the parts move, and the motion rewrites the parts. On \hart it lowers every articulation error by a wide margin and recovers the finger chains of multi-finger hands. Many remaining failures come from the part prior, which fuses parts or breaks a body into pieces; a part prior conditioned on the motion, and estimating the number of links, are left for future work.

\textbf{Limitations.} We require the camera poses to be known and the number of links to be given. A link largely occluded in the first frame is harder to place. Spherical or symmetric parts can spin without change in depth (Fig.~\ref{fig:failure} in the appendix); a color test helps where the texture allows.

{
    \small
    \bibliographystyle{ieeenat_fullname}
    \bibliography{main}
}
\clearpage
\appendix
\setcounter{figure}{0}
\setcounter{table}{0}
\renewcommand{\thefigure}{\Alph{figure}}
\renewcommand{\thetable}{\Alph{table}}
\renewcommand{\theHfigure}{app.\arabic{figure}}
\renewcommand{\theHtable}{app.\arabic{table}}
\twocolumn[{\centering\Large\textbf{Appendix}\par\vspace{1.2em}}]

\section{Method Details}

This section gives the parts of the method that Sec.~\ref{sec:method} summarizes.

\textbf{Placement.} Eq.~\eqref{ak:eq:alignment} is minimized by a two-sided ICP that alternates the correspondences of Eq.~\eqref{ak:eq:cdvis}, every observed point to its nearest generated point and every visible generated point to its nearest observed point, with a weighted closed-form similarity fit, the two directions weighted equally. Since the generator's frame is unrelated to the camera's, the ICP starts from every rotation of the cube's symmetry group, 24 orientations, combined with turns about the vertical, and the best starts are refined further. Each part is then moved rigidly on its own until the start state agrees with $D_0$, by the articulation fit of Sec.~\ref{sec:artic} run against the first frame alone. Where the camera saw the object, the observed surface then replaces the generated one, and the generated surface fills in the rest.

\textbf{Rigid trajectories.} A track $n$ lifted by Eq.~\eqref{ak:eq:lift} gives points $x_{n,f}$ on its valid frames $\mathcal F_n$. A rigid trajectory is a pose $H_{b,f}\in\akSE$ per frame with $H_{b,0}=I$, and it explains a track by the residual
\begin{equation}
 \bar r_{nb}=\frac{1}{|\mathcal F_n|}\sum_{f\in\mathcal F_n}\frac{\|x_{n,f}-H_{b,f}\,x_{n,0}\|^2}{L^2}.
\end{equation}
Sequential RANSAC proposes trajectories from small samples of tracks and keeps the one the most tracks follow, every track is assigned to the trajectory that explains it best, each trajectory is refitted frame by frame to its tracks, and each body is searched again for tracks that move on their own relative to it, so a fingertip turning slightly on its middle link becomes a body of its own.

\textbf{Robust rigid trajectories.} The memberships of Eq.~\eqref{ak:eq:membership} are normalized over the $B$ bodies and the outlier $\varnothing$. Robust track cleaning leaves out every track with $\gamma_{n\varnothing}>\tfrac12$. With robust trajectories, a part's tracked motion from the first to the last frame, which enters the joint search below, is the weighted Procrustes solution
\begin{equation}
 (R^\star,t^\star)=\argmin_{R\in\akSO,\,t}\ \sum_n \gamma_{nb}\,\bigl\|R\,x_{n,0}+t-x_{n,N}\bigr\|^2,
 \label{ak:eq:procrustes}
\end{equation}
alternated with the memberships $\gamma_{nb}$ of Eq.~\eqref{ak:eq:membership} (the part against the outlier) for three rounds from the RANSAC start. Replacing the robust fits by as many fits with new seeds raises the axis error from $10.08^\circ$ to $11.80^\circ$ (Table~\ref{tab:ablation}).

\textbf{Split and merge.} The generated parts are split and merged into links by two kinds of motion evidence. Where the tracks are grouped into bodies, every generated point takes the body of the tracks that start near it, a part that two bodies share is cut along them, parts that move with one body are merged, touching pairs first, and the separate pieces of one body stay separate links, since two fingers curling alike are two links. Where the joints are searched against the frames, the energy $E$ of Sec.~\ref{sec:artic} decides. A connected region of a part whose depth or tracks contradict the part's fitted motion, or a part whose own joint on top of its group's explains the frames better, is split off when that lowers the energy by a margin, $E'\le(1-\delta_s)E$. Two touching groups, or a group and the base, are merged when this raises the energy by little, $E'\le(1+\delta_m)E$, cheapest merge first and, whenever the part count is given, only while more links than that remain, unless ten or more tracks follow the two groups apart. The thresholds are given in Sec.~\ref{sec:supp_impl}.

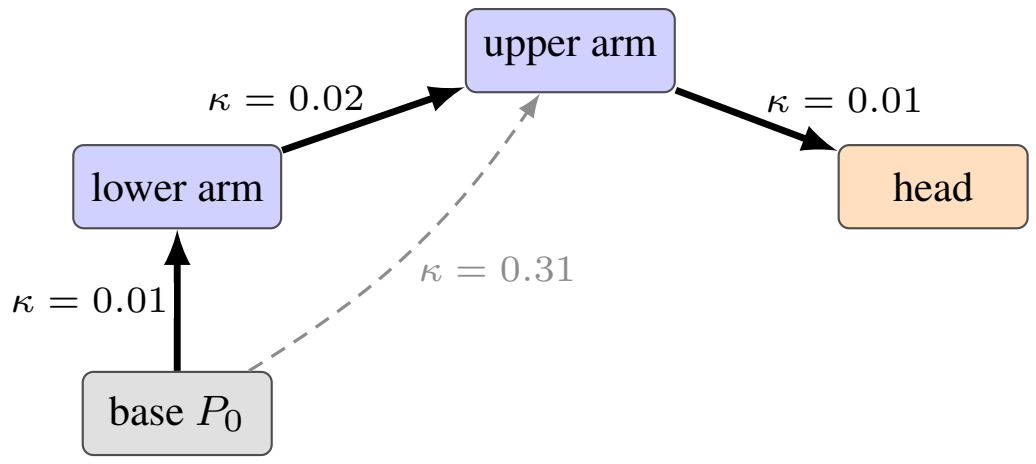
\begin{figure}[t]
\centering
\begin{tikzpicture}[x=1cm,y=1cm,
  lk/.style={draw=black!70,fill=#1,rounded corners=2pt,minimum width=1.25cm,minimum height=0.55cm,font=\footnotesize},
  ch/.style={-latex,line width=1.3pt,black},
  al/.style={-latex,line width=0.6pt,densely dashed,black!45},
  cst/.style={font=\scriptsize,fill=white,inner sep=1pt}]
\node[lk=black!12] (b) at (0,0) {base $P_0$};
\node[lk=blue!18] (a) at (0,1.5) {lower arm};
\node[lk=blue!18] (u) at (2.6,2.4) {upper arm};
\node[lk=orange!25] (h) at (5.0,1.5) {head};
\draw[ch] (b) -- node[cst,left] {$\kappa=0.01$} (a);
\draw[ch] (a) -- node[cst,above left] {$\kappa=0.02$} (u);
\draw[ch] (u) -- node[cst,above right] {$\kappa=0.01$} (h);
\draw[al] (b) to[bend right=12] node[cst,below right] {$\kappa=0.31$} (u);
\end{tikzpicture}
\caption{\textbf{Choosing the tree}, illustrated on a lamp (costs are illustrative). Each arrow joins two touching links and carries the cost $\kappa$ of Eq.~\eqref{ak:eq:treecost} of hanging the child on the parent, which is low when one joint explains their relative motion. The upper arm turns about one fixed axis relative to the lower arm, so that edge is cheap, but relative to the base its motion also contains the lower arm's turn and fits no single joint. The tree is the minimum-cost arborescence rooted at the base (solid).}
\label{fig:tree}
\end{figure}

\begin{figure}[t]
\centering
\includegraphics[width=\linewidth]{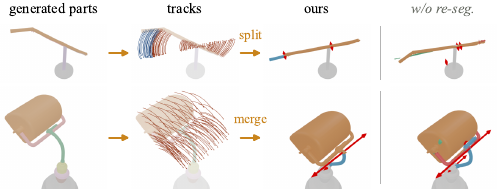}
\caption{\textbf{Motion-driven re-segmentation.} Tracks colored by rigid body. Top: one generated arm moves as two bodies and is split. Bottom: shade and bracket move as one body and are merged. Without re-segmentation both fail at the end state.}
\label{fig:reseg}
\end{figure}

\textbf{The articulation graph.} The parents are the minimum-cost spanning arborescence of the touching links, the tree of directed edges from the base that reaches every link once and has the least total cost. Hanging link $k$ on $p$ costs how far their relative motion $\Delta_{pk,f}=T_{p,f}^{-1}T_{k,f}$ is from a single joint that sits at their contact,
\begin{equation}
 \kappa(p,k)=\frac{\min_{x\in P_k\cap\mathcal N(P_p)}d\bigl(x,\ell_{pk}\bigr)}{L}+\lambda_g\,\frac{h_{pk}}{L}+\lambda_s\Bigl(1-\frac{|P_p|}{|P|}\Bigr),
 \label{ak:eq:treecost}
\end{equation}
where $\ell_{pk}$ is the axis line of the screw of $\Delta_{pk,N}$, $\mathcal N(P_p)$ the points within $1.5\varepsilon$ of the parent, $|P|$ the number of model points, $h_{pk}$ the mean displacement of $P_k$ under $\Delta_{pk,N}$, and the last two terms break ties toward the parent the link moves least from and toward the larger parent. The first term is zero for a relative motion under $\varepsilon$ or without rotation. With trajectories over every frame, the first term becomes the residual of the best one-joint fit of $\Delta_{pk,f}$ over the frames both links cover, since a child turns about one fixed axis relative to its true parent through every frame and in general not relative to any other link, and a constant is added for any parent but the base, since fingers curling together also look like a chain of each other. Fig.~\ref{fig:tree} illustrates the tree selection. Fig.~\ref{fig:reseg} shows the effect on a lamp whose arm was generated as one part and on a lamp whose shade and bracket were generated apart. The first is split along its tracks and the second is merged, and both then articulate correctly.

\textbf{Energy terms.} In Eq.~\eqref{ak:eq:energy} the track term charges every track at every frame it is alive for its distance from where its link carries it,
\begin{equation}
 E_{\mathrm{track}}=\frac{1}{|\mathcal A|}\sum_{(n,f)\in\mathcal A}\min\Bigl(\bigl\|T_{k_n,f}T_{k_n,s_n}^{-1}x_{n,s_n}-x_{n,f}\bigr\|,\,c_{\mathrm{tr}}\Bigr),
\end{equation}
where $k_n$ is the link the track starts on, $s_n$ its first frame, $\mathcal A$ the alive track frames and $c_{\mathrm{tr}}$ a cap. The smoothness term penalizes the second differences of every moving link's amount over the frames,
\begin{equation}
 E_{\mathrm{smooth}}=\frac{L}{|\mathcal K|}\sum_{k\in\mathcal K}\sum_{f=1}^{N-1}\bigl(\bar q_{k,f+1}-2\bar q_{k,f}+\bar q_{k,f-1}\bigr)^2,
\end{equation}
with $\bar q$ an angle in radians or a slide divided by $L$, and $E_{\mathrm{pen}}$ is the fraction of moving points that end deeper than $0.03L$ inside another link's solid, so a lid explains the depth from its own place on the body.

\textbf{Typed joints.} The joint of link $k$ relative to its parent is the screw of Eq.~\eqref{ak:eq:fk}, fitted to the relative motion $\Delta_{pk,f}$. Hypotheses are hinges about the edges and symmetry axes of the link's bounding box, through its center and through its contact with the parent, slides along the box axes, and the joint of the link's tracked motion (Eq.~\eqref{ak:eq:procrustes}). Each is scored by $E$ and the best is refined with its axis held. A freely fitted rigid motion, read back as a joint about its own screw axis, replaces the snapped axis only when it lowers $E$ by more than 5\%. A slide is kept only when it explains the motion about as well as a hinge and barely turns, so a fingertip's small curl stays a hinge, and a link that no track reaches is hinged at its contact with its parent.

\textbf{Two families of fits.} A synchronous group fit ties every joint to one progress through the clip, $q_{k,f}=\theta_f\,q_k$ with $\theta_f=f/N$, so the whole video is explained by one progress shared by all joints and a final amount per joint. The per-frame fit lets every joint move at its own pace. A per-frame chain fit segments the CoTracker3 tracks by their motion into rigid bodies and leaves every $q_{k,f}$ free, which is needed for deep chains of small, self-occluding links such as fingers, since every frame's depth and the long tracks then determine each joint's amount. There, frames where a link's tracks cover it take the amount from its trajectory, and frames its tracks have lost take it from a line search of $q_{k,f}$ against that frame's depth, in tree order with the link's descendants riding along. All joints are then refined together by gradient descent on Eq.~\eqref{ak:eq:energy} over the axes $(a_k,o_k)$ and the amounts, the tree is estimated again from the poses now complete over every frame, and the chain refitted when it changes. The synchronous fit carries the penetration term and the per-frame fit the smoothness term; the weights are given in Sec.~\ref{sec:supp_impl}.

\section{Implementation Details}
\label{sec:supp_impl}

\textbf{Generative part prior.} The start image is cropped to a square around the object mask with the object filling 85\% of the side and resized to $518\times518$. Hunyuan3D-2.1~\cite{hunyuan3d21} generates the complete mesh from the crop, P3-SAM~\cite{p3sam} proposes part labels on it (on the hands also a finer run with 1{,}000 prompts), and X-Part~\cite{xpart} regenerates each part as a complete mesh.
The pools of the hands include this finer decomposition, and those of the objects use P3-SAM's default prompting, which gives one to twenty parts per object. The generated parts are sampled to 30{,}000 points in total, each in proportion to its area, and each part separately to 3{,}000 points for the local energies.

\textbf{Placement and start correction.} The similarity transform from the generated to the observed frame is initialized at the 24 axis-aligned orientations combined with 12 turns about the vertical axis, each refined by six iterations of a two-sided ICP on 6{,}000 generated and 1{,}500 observed points, and the eight best candidates are refined for 30 iterations on 20{,}000 and 6{,}000 points. The start correction then runs the articulation fit of Sec.~\ref{sec:artic} against the start view alone, so that each generated part is moved rigidly to where the image shows it. Observed points within $3\varepsilon$ of a generated part take its label, generated points further than $2\varepsilon$ from every observed point are kept as the unseen geometry, and everything else of the generated surface is replaced by the observed one, with $\varepsilon=0.02$ of the object's diagonal.

\textbf{Visibility-aware energy.} A posed model is scored against a frame by the mean capped distance from the observed surface to the posed model and from the visible model points to the observed surface, with the cap at 0.10 of the diagonal; a model point is visible when no model point lies more than $\varepsilon$ in front of it along its ray, read from a point z-buffer of the posed model itself. Interpenetration adds the cap times the fraction of a part's points that lie deeper than 0.03 of the diagonal (at least two voxels of an 80-voxel grid) inside another group's solid, which is the dilated and hole-filled voxelization of the group at the start. A merge may raise the energy by at most 1\% of the fitted value. The energy weighs the end frame by 1 and ten evenly spread in-between frames by 0.3 each, whereas the joint search uses five of these frames at 0.6 each.

\textbf{Joint search.} Candidate revolute axes are the twelve edge directions and three symmetry axes of a part's oriented bounding box through its center and through the contact line with its parent, plus the axis of the part's own rigid motion between frames; candidate prismatic axes are the three box axes and the motion direction. Each candidate is screened at turns of $20^\circ$ (or the matching slides) and refined by ICP with the angle free. A fit leaves the snapped axis set only if the free axis lowers the energy by more than 5\%, and among fits within 10\% of the best gain the smallest motion is kept. A part left in place is searched again later in the video if 10\% of its visible points look off, and that search is kept if it lowers the energy by 1\%. The path of every motion is checked for interpenetration at a quarter, half and three quarters of the way.

\textbf{Kinematic tree.} The parents are the minimum-cost spanning arborescence (Edmonds' algorithm) of Eq.~\eqref{ak:eq:treecost} over the touching pairs, with $\lambda_g=\lambda_s=0.01$; a link that touches nothing can hang on the base at cost 10. Two conventions follow the PartNet-Mobility trees where any tree explains the motion equally. A link hung on a smaller link that moves with it swaps places with it, and links turning about one hinge line relative to a part they all touch hang on the largest such part. On the per-frame fits the edge cost is the residual of the one-joint fit of the relative trajectory over the frames both links cover, the relative motion at the last common frame, the gap between the two links beyond $1.5\varepsilon$, and a constant for any parent but the base.

\textbf{Tracks.} Every object pixel of the first frame on a stride of two is followed frame to frame by RAFT optical flow~\cite{raft} (the large model, chained), ends where the forward and backward flows disagree by more than 1.5 pixels or the point leaves the object mask, and is followed back to the first frame by the backward flow; tracks whose round trip ends more than a few percent of the diagonal from their start are dropped. A track is lifted to 3D by the depth of each frame, and a part's motion hypothesis is scored on how far it carries the part's tracks. Color is not a term of $E$ or $S$. Instead, it enters the group fits through a single test after the fit, in which a textured part is turned about its own symmetry axes; we keep a turn when it lowers the misfit of the end frame's colors by 20\% while $E$ stays within 1\%.

\textbf{Re-segmentation.} A contradicted region is a connected set of at least 40 points covering 5\% to 70\% of a part whose observed depth or tracks contradict the part's fitted motion. The three largest regions per round are split off, each split is kept if it lowers the energy by 2\%, keeps half of the new part's visible points visible and lowers the part's local energy by 10\%, and the points between the two pieces are reassigned for six rounds. One round of re-segmentation is run on each decomposition, followed by two rounds of relabeling points to the part whose motion the frames support.

\textbf{The pool of fits.} The objects receive the pool on a single part decomposition, whereas the hands receive the group family on both of their decompositions and the chain family on the finer one. The group family holds twelve synchronous fits. Eight run with the track cap $c_{\mathrm{tr}}$ and the track preference perturbed (caps of $0.10L$ and $0.04L$, the tracked motion preferred or not, two seeds). Two use robust track cleaning and leave out the tracks that no body explains ($\gamma_{n\varnothing}>\tfrac12$ with bodies from sequential RANSAC at a mean distance of $0.015L$ and inlier radius $\rho=0.015L$). Two use robust trajectories and estimate every part's tracked motion by three rounds of Eq.~\eqref{ak:eq:procrustes} with $\rho=2\varepsilon$, each pair at two of the perturbations. The chain family holds nine per-frame fits on CoTracker3 tracks. Five of them read the links in different ways (tracks grouped into bodies, dense trajectories of the generated shape, or cuts along the kinematic chain) or under different motion segmentation settings, two use robust track cleaning, and two use robust trajectories over every frame, both with $\rho=0.015L$. On the hands, the chain family holds twenty-two fits, with further variants of the same kinds. Selection follows Eq.~\eqref{ak:eq:select}. The group fits of one decomposition share one parametrization, and the lowest $E$ among them is that decomposition's finalist; the chain fit with the lowest check $S$ (Eq.~\eqref{ak:eq:check}) is the chain finalist; and the finalist with the lowest $S$ is kept. For the synchronous fits $E$ weighs the end frame by 1 and ten evenly spread in-between frames by 0.3 each ($\lambda_{\mathrm{pen}}=\lambda_{\mathrm{tr}}=1$, $\lambda_{\mathrm{sm}}=0$), and the joint search ranks hypotheses on five of them; for the per-frame fits every frame weighs $1/(N+1)$, with $\lambda_{\mathrm{tr}}=1$, $c_{\mathrm{tr}}=0.1L$, $\lambda_{\mathrm{sm}}=0.3$, $\lambda_{\mathrm{pen}}=0$ and 400 Adam steps at a learning rate of 0.01. Splits are kept at $\delta_s=0.02$ and merges at $\delta_m=0.01$.

\textbf{Link re-generation.} Every link of the selected fit re-prompts X-Part, on every input. The Chamfer distances of Eq.~\eqref{ak:eq:linkpick} are computed on 4{,}000 points sampled from each side. The joints and poses of the selected fit are kept. After the choice, the observed surface of each link, fused over the frames, replaces the part of the generated surface that the camera sees.

\textbf{The chain fits.} The chain fits use every frame and fit the articulation over all 30 frames. Tracks come from CoTracker3~\cite{cotracker3} (the offline model), every object pixel of the first frame on a stride of three followed through the video and, from the last frame, back through it; a point counts at every frame the tracker marks it visible and on the object mask, so a fingertip that curls out of sight is picked up again where it re-emerges. The tracks are grouped into rigid bodies over the video and the generated parts cut and merged along them to the part count as in the main method, with two rules added. The separate pieces of a body are separate links (fingers curling alike read as one body), and the point labels are made spatially coherent (majority over the sixteen nearest points, pieces under 80 points joining their neighbors). Links whose motions nearly agree after the fit are not merged. The links' tree is the cheapest arborescence from the static base where hanging a link on a parent costs the residual of a one-joint fit to their relative trajectory over the frames both cover, the relative motion at the last common frame and the gap between them, plus a constant for any parent but the base. Each joint's axis and type come from that one-joint fit; its amount at every frame comes from the trajectories where the link's tracks cover the frame and from a line search against the frame's depth elsewhere (the descendants riding along, in tree order). All joints are then refined together by gradient descent for 400 steps on the visibility-aware Chamfer of every frame against its depth, every track at every frame it is visible (carried from its first frame by its link's pose), and a smoothness prior on the amounts; the tree is re-estimated from the fitted trajectories and the fit repeated once if it changed. The fit returns a pose per link per frame, so each frame's depth is labeled by the nearest posed link. Each link's observed surface over the video, brought back to the start pose by its poses and fused on a voxel grid, replaces the generated surface wherever the video saw it (the generated points fill in the rest).

\textbf{Run time.} Per object, generating the shape and its parts takes about 8\,min on one A800 GPU, the tracks about 2\,min, the pool of fits about 20\,min on CPU with the fits run in parallel, and the selection with link re-generation about 10\,min, about 40\,min in all.

\section{Benchmark Construction}

\textbf{\harto.} Each PartNet-Mobility object is rendered with Blender from 100 training and 20 test cameras at $800\times800$ in its rest state and in an end state in which every joint has moved by a random fraction of its range, with per-view RGB, depth, object mask and part-label maps. The input view is one of the training cameras, and a 30-frame video is rendered from that camera with every joint interpolated from rest to the end state. \harto holds 60 objects of eight categories, each with two to ten joints and at least one joint chain (85 of the 159 joints hang on a moving link): 27 lamps, 19 staplers, 6 globes, 3 folding chairs, 2 safes, and one box, knife and pliers. Lamps and staplers make up most of \harto since joint chains with thin parts are common among them, while closed furniture with one level of joints is covered by Articulate-100. Ground truth is the posed part mesh at every state and the joint axis, type and motion of every part, from which the world-frame motion of each part is computed by registering its start and end meshes.

\textbf{\harth.} Eight hands and grippers from dex-urdf~\cite{dexurdf,bunnyvisionpro} (sources in Sec.~\ref{sec:setup}) are converted to the same object format. Only the visual meshes are kept, mimic joints become independent joints with the coupling recorded, fixed joints are kept and composed, and each model is scaled uniformly so that its rest bounding-box diagonal matches the household objects. Each hand is rendered open at its URDF rest pose and closed 40 to 60\% of the way to its joint limits, with the same cameras and video as \harto. The joints whose child link carries a mesh are scored, 93 in all.

\section{Protocol and Baselines}
\label{sec:supp_protocol}

\textbf{Matching.} Each method's parts are written as an articulated model with one part per node. Predicted and ground-truth parts are paired by a Hungarian assignment on the Chamfer distance between every predicted and every true part, averaged over the start and end states, and the same pairing serves the geometry and the joints, so each part is scored for shape and motion against the same ground-truth part. Every ground-truth joint is scored through the part paired with its link. The motion of a part, true or predicted, is taken relative to the part paired with its ground-truth parent, so a link in a chain is scored on its own joint and not on the motion it inherits. The predicted motion of that part is read as it is, as a hinge about its rotation axis when it turns by $5^\circ$ or more and as a slide along its translation otherwise. The axis angle is the angle between the predicted and true axis directions, the axis position the shortest distance between the two axis lines in units of 0.1\,m, measured on revolute joints only, and the part motion the angle (revolute) or distance (prismatic) between the predicted and true motion of the part from the start to the end state. A predicted type that differs from the truth keeps its axis angle and receives the motion penalty, and on a true hinge the axis position penalty. An unpaired joint, a paired part that stays static, or a failed object receives $90^\circ$, 10, and $90^\circ$ or 1\,m. Chamfer distances are measured in mm on the static part and on each movable part, averaged over the start and end states, with 1000\,mm for an unpaired part.

\textbf{AutoURDF and REART} receive point cloud sequences sampled from the posed ground-truth meshes along the motion, independent of the 30 video frames, with the ground-truth part count. REART receives ten complete posed meshes at evenly spaced joint values, with 4{,}096 points per frame. AutoURDF receives the input of its own protocol, namely five sequences of ten frames, one of which moves every joint from the start to the end state while the other four are random walks within the joint ranges; each frame is fused from the depth of 20 cameras above the object and sampled to 5{,}000 points. Both run with their released code. However, REART's part count is fixed to $K$ with part merging turned off, and AutoURDF is given the ground-truth number of joints instead of estimating it. \textbf{Articulate-Anything} receives the 30-frame video and runs in its video modality with GPT-4o and all-joints mode, with the tested object removed from its retrieval library. Its template is rotated into the ground-truth frame using the base orientation of the ground-truth URDF, or the best of the 24 axis-aligned rotations where that fails, then placed and scaled onto the bounding box of the ground-truth start mesh, and finally scored by the same protocol.

\textbf{Real captures.} We use the first take of each of RBO's fourteen objects and a second take of the globe, the laptop and the pliers, which gives 17 sequences with 22 joints. Each sequence is cut to 30 frames spread evenly from the first frame to the frame of largest joint motion, and the ground-truth part meshes are RBO's object models posed by the motion capture. CD-w denotes the Chamfer distance of the whole object. On RBO the object mask of every frame is the set of pixels whose depth lies within 1.5\,cm of the object models posed by the motion capture, and on iTACO it is the SAM~3 mask of the object's name with the hands removed. Every method receives observations from the same recordings, with the same masks. AutoURDF and REART receive the object's masked depth of 10 of the 30 frames, lifted with each frame's camera, and Articulate-Anything the 30-frame video; the part count is given to AutoURDF and REART. \ours runs as on \hart with three additions that apply to every object alike. Hands and arms cover the object in many frames, so each sequence is generated once, from one of three inputs, the one whose fit the video check of Eq.~\eqref{ak:eq:select} preferred in a first pass with an earlier set of fits: the first frame, an automatic edit of the first frame by Qwen-Image-Edit-2509~\cite{qwenimage} that removes the person, the table, the background and the motion-capture markers and completes the object, with one fixed prompt in which only the object's name changes, or the last frame with the clip played backwards. The pool and the selection are those of \harto.
Pixels of the person, found by a person segmenter, are treated as occluders by the energy and are left out of the chain fits' tracks.
The selection of Eq.~\eqref{ak:eq:select} runs over the fits of that one generation, a fit from the reversed clip being checked against the forward clip; on the real captures only fits with $K$ links compete. Articulate-Anything's retrieved template comes in its own frame. As on \hart, it is placed with the help of the ground truth: it is kept upright, turned about the vertical to best match the ground-truth start and end meshes, with the template opened by its own predicted joints for the end state, and scaled to fit the bounding box of the ground-truth start mesh. Scoring follows the protocol above, and a matched part that does not move receives the penalty. All 22 joints are scored, with the axis position measured on the 15 revolute ones.

\textbf{Generated video.} We make the start and end key frames of the excavator with Qwen-Image-Edit-2509~\cite{qwenimage}, and the first-last-frame model of Wan2.1 (14B)~\cite{wan2025} generates 81 frames between them at $832\times480$, prompted to unfold the boom, the stick and the bucket together in front of a static camera. The clip is then stabilized on the base and resampled to 30 frames spread evenly in motion. Depth Anything 3~\cite{depthanything3} provides the depth and the camera, SAM~3~\cite{sam3} masks the object from its name, and $K=4$ (base, boom, stick and bucket). \ours runs as on the real captures, while REART and AutoURDF receive the masked depth of 10 of the 30 frames.

\section{Additional Results}

\textbf{Part quality of link re-generation.} The Chamfer distance barely changes with link re-generation, since a link assembled from several generated parts covers the same surface as one regenerated part. Table~\ref{tab:supp_regen_quality} measures the links themselves. Keep-closer re-generation replaces 57\% of the moving links on \harto, and the replaced links have fewer components and more often come from one generated part. Link re-generation thus makes the parts more coherent, with fewer pieces per link, rather than lowering the Chamfer distance.

\begin{table}[!t]
\centering
\caption{\textbf{Part quality of link re-generation} on the moving links of \harto (60 objects), with and without re-generation, for the same selected fits. Mean over objects, with the paired 95\% bootstrap interval of the difference. A component is a set of faces joined by shared vertices; a source part is a generated part that covers at least 5\% of the link's area.}
\label{tab:supp_regen_quality}
\setlength{\tabcolsep}{3pt}
\resizebox{\linewidth}{!}{
\begin{tabular}{lccc}
\toprule
 & without & full & difference \\
\midrule
Components per link $\downarrow$ & 3.33 & \textbf{2.84} & $-0.49$ [$-0.70$, $-0.30$] \\
Links in one component (\%) $\uparrow$ & 22.5 & \textbf{32.9} & $+10.4$ [$+3.8$, $+17.8$] \\
Links from one source part (\%) $\uparrow$ & 39.4 & \textbf{50.9} & $+11.5$ [$+4.4$, $+19.2$] \\
Source parts per link $\downarrow$ & 2.11 & \textbf{1.85} & $-0.27$ [$-0.41$, $-0.14$] \\
\bottomrule
\end{tabular}}
\end{table}

\textbf{Per hand.} Table~\ref{tab:supp_hands} lists \ours per hand under the protocol of Table~\ref{tab:main}, with and without link re-generation. Every scored joint is paired with a link on every hand. The selection keeps a chain fit on six hands and a group fit on the Barrett hand and the Panda gripper.

\begin{table}[!t]
\centering
\caption{\textbf{\ours per hand on \harth}, with link re-generation (full) and without it: axis angle ($^\circ$), part motion ($^\circ$) and CD-s / CD-m (mm).}
\label{tab:supp_hands}
\setlength{\tabcolsep}{2.5pt}
\resizebox{\linewidth}{!}{
\begin{tabular}{lcccccccc}
\toprule
 &  & \multicolumn{3}{c}{full} & \multicolumn{3}{c}{without re-generation} \\
\cmidrule(lr){3-5} \cmidrule(lr){6-8}
Hand & Joints & Axis & Motion & CD-s / CD-m & Axis & Motion & CD-s / CD-m \\
\midrule
Allegro & 16 & 31.3 & 63.9 & 90.3 / 198.0 & 31.3 & 63.9 & 90.3 / 198.6 \\
Schunk & 20 & 41.9 & 47.5 & 38.8 / 47.1 & 41.9 & 47.5 & 38.8 / 47.4 \\
Ability & 10 & 30.6 & 53.3 & 38.8 / 77.1 & 30.6 & 53.3 & 38.8 / 78.1 \\
LEAP & 16 & 30.9 & 36.4 & 23.5 / 62.1 & 30.9 & 36.4 & 23.5 / 62.1 \\
D'Claw & 9 & 4.6 & 3.7 & 12.1 / 13.3 & 4.6 & 3.7 & 12.1 / 13.3 \\
Barrett & 8 & 20.9 & 20.7 & 14.0 / 15.9 & 20.9 & 20.7 & 14.0 / 15.5 \\
Inspire & 12 & 37.1 & 40.6 & 21.4 / 72.4 & 37.1 & 40.6 & 21.4 / 73.5 \\
Panda & 2 & 0.0 & 0.0 & 23.5 / 12.5 & 0.0 & 0.0 & 23.5 / 12.8 \\
\bottomrule
\end{tabular}}
\end{table}

\textbf{Selection rule.} The selection of Eq.~\eqref{ak:eq:select} picks a chain fit on 16 of the 60 \harto objects and on 6 of the 8 hands. We compared it with three other GT-free rules fixed before scoring, each over the same pool and with the same link re-generation. Ranking every fit by $S$ alone gives the same picks on the hands and a higher axis error on \harto ($10.04^\circ$ against $8.13^\circ$). Ranking every fit by $E$ alone ($10.68^\circ$ on \harto, $28.31^\circ$ on \harth) and by the sum of the ranks under $E$ and $S$ ($11.39^\circ$ and $25.49^\circ$) are worse on both. Thus, $E$ compares only group fits that share their links, whereas $S$ compares all other fits.

\textbf{Real captures.} Table~\ref{tab:supp_real} compares the methods on RBO.
\ours has the lowest errors apart from the axis position, where AutoURDF is lower, and roughly halves the Chamfer distance of the movable parts against the best baseline. Symmetric parts such as the globes and the Rubik's cube turn without changing the depth, so their axes are poorly determined.

\begin{table}[t]
\centering
\caption{\textbf{Comparison on RBO}~\cite{rbo} (17 sequences, 22 joints, motion-capture ground truth). Articulation: axis angle ($^\circ$), axis position (0.1\,m), part motion ($^\circ$ or m). Geometry: Chamfer distance of the movable parts and of the whole object (mm). Parts are paired and joints scored as in Table~\ref{tab:main} of the main paper; unpaired joints and failed objects receive the penalty. Best in \textbf{bold}, second \underline{underlined}. Gain is the improvement over the best baseline in each column.}
\label{tab:supp_real}
\setlength{\tabcolsep}{3pt}
\resizebox{\linewidth}{!}{
\begin{tabular}{lcccccc}
\toprule
\multirow{2}{*}{Method} & \multicolumn{3}{c}{Articulation} & \multicolumn{2}{c}{Geometry} & \multirow{2}{*}{Fail $\downarrow$} \\
\cmidrule(lr){2-4} \cmidrule(lr){5-6}
 & Axis Ang $\downarrow$ & Axis Pos $\downarrow$ & Part Motion $\downarrow$ & CD-m $\downarrow$ & CD-w $\downarrow$ & \\
\midrule
AutoURDF~\cite{autourdf} & 51.05 & \textbf{1.39} & 98.40 & \underline{80.4} & 45.8 & 0/17 \\
REART~\cite{reart} & 57.62 & 8.19 & 86.93 & 479.9 & 434.2 & 7/17 \\
Articulate-Anything~\cite{articulate-anything} & \underline{40.33} & 2.94 & \underline{75.26} & 121.6 & \underline{42.7} & 0/17 \\
\rowcolor{ourscolor} \ours & \textbf{32.45} & \underline{2.11} & \textbf{68.36} & \textbf{41.0} & \textbf{25.2} & 0/17 \\
\midrule
\rowcolor{gray!12} Gain & +7.88 & $-$0.72 & +6.90 & +39.4 & +17.5 & \\
\bottomrule
\end{tabular}}
\end{table}

\begin{figure*}[t]
\centering
\includegraphics[width=\textwidth]{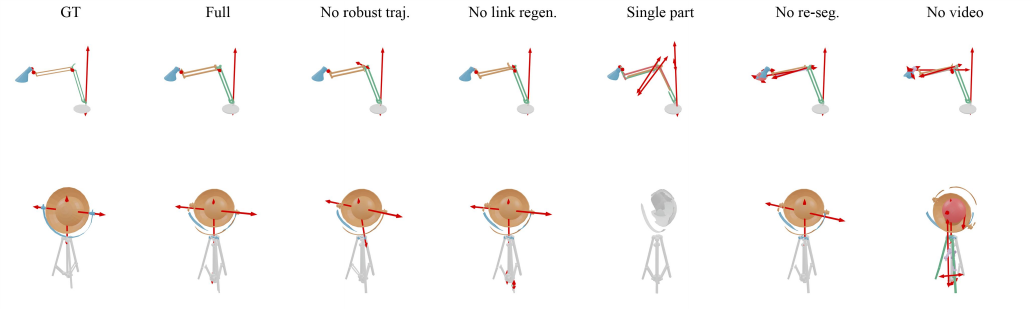}
\caption{\textbf{Ablations at the end state} on two lamps, parts colored by the ground-truth part they are paired with, showing the ground truth, the full model and single-component ablations.}
\label{fig:ablation_qual}
\end{figure*}

Fig.~\ref{fig:ablation_qual} shows single-component ablations at the end state on two lamps. On the first lamp, without robust trajectories the axis error grows from $3.9^\circ$ to $12.9^\circ$, and the single generated part, the decomposition without re-segmentation and the fit without video each tilt the axis by over $30^\circ$. On the second lamp the single generated part gives an axis error of $71.2^\circ$ and a CD-m of 520 against 37\,mm, and the fit without video tilts the axis by $44.3^\circ$.

\begin{figure*}[t]
\centering
\includegraphics[width=0.85\textwidth]{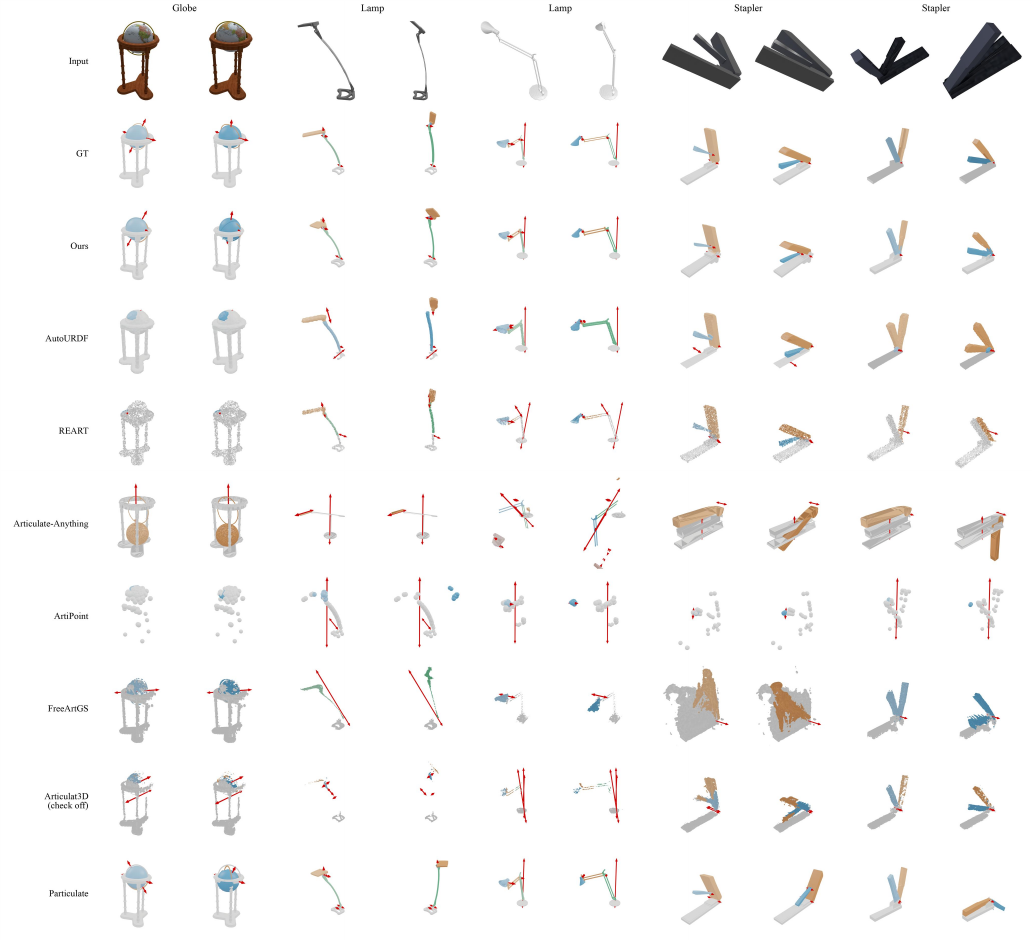}
\caption{\textbf{More results on \harto}: input, ground truth, \ours, the baselines of Table~\ref{tab:main} and the further baselines of Sec.~\ref{sec:supp_more} at the start and end state, in the ground truth's frame, with joint axes in red. Gray marks each method's own root part. Articulat3D is the variant with its own joint types and its check off. Particulate~\cite{particulate} runs on the geometry and placement of \ours, and its end state is its predicted joint limit.}
\label{fig:supp_qual}
\end{figure*}

\textbf{Additional qualitative results.} Fig.~\ref{fig:supp_qual} shows five more \harto objects with every baseline, including the further baselines of Sec.~\ref{sec:supp_more}.

\section{Post-hoc Texture}

\begin{figure}[t]
\centering
\includegraphics[width=\linewidth]{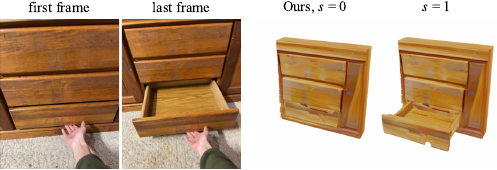}
\caption{\textbf{Texture added after the fit.} The iTACO dresser, first and last input frame, then the textured \ours model at the start ($s=0$) and end ($s=1$) state, seen from the first frame's camera.}
\label{fig:texture}
\end{figure}

Texture is not the focus of this work, and the parts are scored and shown by their geometry. Since every part comes from a generated mesh, a texture can be added afterwards. Hunyuan3D-2.1 Paint~\cite{hunyuan3d21} paints the generated whole mesh from the input image, and each face of the articulated parts takes the color of the nearest painted surface point. The parts then move with their texture (Fig.~\ref{fig:texture}): the dresser's drawer slides out with its wood grain and its inner floor, which the input image never shows.

\section{Failure Cases}

\begin{figure}[t]
\centering
\includegraphics[width=\linewidth]{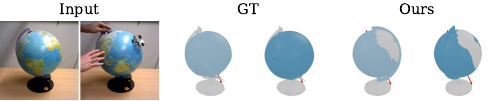}
\caption{\textbf{Failure cases.} Input, ground truth and \ours at the start and end state. On the first RBO globe the sphere spins about its own axis, which leaves the depth unchanged. The recovered axis is close to the ground truth, but the meridian ring is fused into the spinning sphere and a patch of the sphere stays with the stand.
}
\label{fig:failure}
\end{figure}

The failures have three sources. First, a part that spins or slides along its own symmetry leaves the depth unchanged (Fig.~\ref{fig:failure}), and the tracks, together with the color test of the group fits, resolve it where the texture allows. A generated part that fuses two true parts is split where the video moves the halves apart, so a joint the video exercises little stays fused. And a generated part posed far from the image, a lid generated open, is corrected rigidly at the start, and a large enough correction moves the part off the observed surface.

\section{Additional Baselines and Analyses}\label{sec:supp_more}
\renewcommand{\topfraction}{0.9}\renewcommand{\textfraction}{0.05}\renewcommand{\floatpagefraction}{0.85}\renewcommand{\dblfloatpagefraction}{0.85}\setcounter{topnumber}{4}\setcounter{dbltopnumber}{3}\setcounter{totalnumber}{6}%
\newsavebox{\syfitbox}\newcommand{\syfit}[1]{\sbox{\syfitbox}{\footnotesize #1}\ifdim\wd\syfitbox>\linewidth\resizebox{\linewidth}{!}{\usebox{\syfitbox}}\else\usebox{\syfitbox}\fi}
\begin{table*}[!t]
\centering
\footnotesize
\caption{\textbf{Further baselines} on \harto (left) and \harth (right), with the protocol and units of Tables~\ref{tab:main} and~\ref{tab:hands}. The first three rows are copied from those tables. Articulat3D (GT priors) receives the ground-truth joint types and parents, as its released code expects. n/a: the method returns no geometry.}
\label{tab:supp_newbl}
\setlength{\tabcolsep}{3.5pt}
\resizebox{\textwidth}{!}{\begin{tabular}{lccccccp{2pt}cccccc}
\toprule
 & \multicolumn{6}{c}{\harto} & & \multicolumn{6}{c}{\harth} \\
\cmidrule(lr){2-7} \cmidrule(lr){9-14}
Method & Axis Ang & Axis Pos & Part Motion & CD-s & CD-m & Fail & & Axis Ang & Axis Pos & Part Motion & CD-s & CD-m & Fail \\
\midrule
AutoURDF~\cite{autourdf} & 25.02 & 3.36 & 28.10 & 186.8 & 227.2 & 7/60 &  & 54.73 & 4.35 & 61.41 & 463.1 & 444.1 & 3/8 \\
REART~\cite{reart} & 29.69 & 3.41 & 35.59 & 106.6 & 241.2 & 0/60 &  & 68.09 & 6.62 & 70.91 & 37.8 & 473.6 & 0/8 \\
Articulate-Anything~\cite{articulate-anything} & 53.37 & 8.32 & 93.72 & 557.4 & 613.6 & 2/60 &  & 79.18 & 9.07 & 86.35 & 733.1 & 786.2 & 2/8 \\
ArtiPoint~\cite{artipoint} & 72.74 & 9.54 & 76.02 & n/a & n/a & 0/60 &  & 81.02 & 9.49 & 82.34 & n/a & n/a & 0/8 \\
FreeArtGS~\cite{freeartgs} & 63.35 & 6.64 & 69.15 & 200.5 & 660.0 & 0/60 &  & 80.24 & 9.31 & 81.21 & 125.4 & 877.5 & 0/8 \\
Articulat3D~\cite{articulat3d} (GT priors) & 81.57 & 9.11 & 82.11 & 875.2 & 893.9 & 52/60 &  & 78.80 & 10.00 & 78.75 & 881.5 & 879.1 & 7/8 \\
Articulat3D~\cite{articulat3d} (own types) & 84.67 & 9.47 & 85.42 & 920.8 & 928.0 & 55/60 &  & 78.80 & 10.00 & 78.75 & 881.6 & 878.5 & 7/8 \\
Articulat3D~\cite{articulat3d} (own types, check off) & 30.17 & 3.89 & 41.50 & 59.6 & 116.5 & 0/60 &  & 41.29 & 3.05 & 47.84 & 229.5 & 203.4 & 1/8 \\
\rowcolor{ourscolor} \ours & \textbf{8.13} & \textbf{0.60} & \textbf{8.47} & \textbf{26.1} & \textbf{45.6} & 0/60 &  & \textbf{24.67} & \textbf{1.01} & \textbf{33.27} & \textbf{32.8} & \textbf{62.3} & 0/8 \\
\bottomrule
\end{tabular}}
\end{table*}
\begin{table}[!t]
\centering
\footnotesize
\caption{\textbf{Kinematic tree.} Share of ground-truth parent edges reproduced (object mean) and share of objects with the correct root, each read from the method's own tree.}
\label{tab:supp_tree}
\setlength{\tabcolsep}{3.5pt}
\syfit{
\begin{tabular}{lcccc}
\toprule
 & \multicolumn{2}{c}{\harto} & \multicolumn{2}{c}{\harth} \\
\cmidrule(lr){2-3} \cmidrule(lr){4-5}
Method & Edges $\uparrow$ & Root $\uparrow$ & Edges $\uparrow$ & Root $\uparrow$ \\
\midrule
AutoURDF~\cite{autourdf} & 0.48 & 0.73 & 0.33 & 0.25 \\
REART~\cite{reart} & 0.33 & 0.67 & 0.32 & 0.88 \\
Articulate-Anything~\cite{articulate-anything} & 0.19 & 0.47 & 0.16 & 0.25 \\
\rowcolor{ourscolor} \ours & \textbf{0.84} & \textbf{0.98} & \textbf{0.77} & \textbf{1.00} \\
\bottomrule
\end{tabular}}
\end{table}
\begin{table}[!t]
\centering
\caption{\textbf{Axis angle ($^\circ$) by structure.} Objects of \harto grouped by number of joints and by tree depth (levels of joints below the root), and joints whose parent is a moving link (chain joints) on \harto and \harth.}
\label{tab:supp_chain}
\setlength{\tabcolsep}{3pt}
\syfit{
\begin{tabular}{lccccc}
\toprule
Group & n & AutoURDF & REART & Art.-Anything & \ours \\
\midrule
\harto, 2 joints & 39 obj. & 18.55 & 18.88 & 44.75 & \textbf{8.35} \\
\harto, 3 or more joints & 21 obj. & 37.05 & 49.75 & 69.37 & \textbf{7.73} \\
\harto, 2 joint levels & 43 obj. & 21.75 & 22.49 & 48.34 & \textbf{7.98} \\
\harto, 3 or more joint levels & 17 obj. & 33.29 & 47.89 & 66.10 & \textbf{8.54} \\
\midrule
\harto, chain joints & 85 joints & 26.30 & 42.11 & 61.88 & \textbf{13.44} \\
\harth, chain joints & 63 joints & 51.68 & 69.96 & 87.64 & \textbf{32.09} \\
\bottomrule
\end{tabular}}
\end{table}
\begin{table}[!t]
\centering
\caption{\textbf{Selected fit against the pool.} Axis angle and part motion ($^\circ$) of the fit chosen by Eq.~\eqref{ak:eq:select}, of a random candidate (expectation) and of the median candidate.}
\label{tab:supp_select}
\setlength{\tabcolsep}{3.5pt}
\syfit{
\begin{tabular}{lcccc}
\toprule
 & \multicolumn{2}{c}{\harto} & \multicolumn{2}{c}{\harth} \\
\cmidrule(lr){2-3} \cmidrule(lr){4-5}
 & Axis & Motion & Axis & Motion \\
\midrule
Random candidate & 17.96 & 23.18 & 37.84 & 50.00 \\
Median candidate & 14.45 & 16.23 & 38.13 & 50.25 \\
\rowcolor{ourscolor} Selected (\ours) & \textbf{8.13} & \textbf{8.47} & \textbf{24.67} & \textbf{33.27} \\
\bottomrule
\end{tabular}}
\end{table}
\begin{table}[!t]
\centering
\caption{\textbf{Joints from shape or from motion}, on the same generated and registered geometry: axis angle ($^\circ$) and axis position (0.1\,m).}
\label{tab:supp_shape}
\setlength{\tabcolsep}{3.5pt}
\syfit{
\begin{tabular}{lcccc}
\toprule
 & \multicolumn{2}{c}{\harto} & \multicolumn{2}{c}{\harth} \\
\cmidrule(lr){2-3} \cmidrule(lr){4-5}
Joints from & Axis Ang & Axis Pos & Axis Ang & Axis Pos \\
\midrule
Shape, Particulate~\cite{particulate} & 9.58 & 1.53 & 82.83 & 8.48 \\
\rowcolor{ourscolor} Motion, \ours & \textbf{8.13} & \textbf{0.60} & \textbf{24.67} & \textbf{1.01} \\
\bottomrule
\end{tabular}}
\end{table}

\textbf{Further baselines.} We also run three recent methods on \hart. ArtiPoint~\cite{artipoint} estimates joints from point tracks in an RGB-D interaction video, FreeArtGS~\cite{freeartgs} reconstructs an articulated object with Gaussian splatting from a monocular RGB-D video, and Articulat3D~\cite{articulat3d} fits articulated Gaussians with typed joints to a monocular video. Each receives the benchmark RGB-D video, the object masks, the camera and the part count, restricted to what its released code consumes, and runs with its released code, checkpoints and default settings. ArtiPoint segments the video by the hand that moves the object, and the benchmark videos show no hand, so the whole clip is one interaction and its query points are drawn inside the object mask. It returns one joint and no geometry, so its Chamfer distances are not reported. FreeArtGS models one moving part, and the remaining joints are unmatched. The released Articulat3D reads ground-truth joint types and parents from the object's annotation; we report it with these, and with them replaced by a split of the $K-1$ joints taken from its own track classifier. Its second stage matches joint candidates to parts by enumerating all permutations, which does not finish on the hands with 16 and 20 joints, so there it uses the Hungarian algorithm, which returns the same assignment. Its built-in check, which stops a run when the articulated model renders more than 5\,dB below its free-motion model, rejects most objects in both variants, and these count as failures. We also run the own-types variant with this check switched off through its configuration value and nothing else changed (check off). Its one failure is a hand where the second stage assigns no Gaussians to one part. Fig.~\ref{fig:supp_qual} shows these methods on five objects. Scoring follows Sec.~\ref{sec:supp_protocol}. Table~\ref{tab:supp_newbl} lists the results next to the rows of Tables~\ref{tab:main} and~\ref{tab:hands}.

\textbf{Kinematic tree.} The tree of each method is read from its own output (the parent of each link for \ours, the joint connections for REART, the URDF for AutoURDF and Articulate-Anything) and mapped to the ground truth through the part pairing of the protocol. Table~\ref{tab:supp_tree} reports the share of ground-truth parent edges that each method reproduces, averaged over objects, and the share of objects whose root is paired with the ground-truth static part. A failed object reproduces no edge.

\textbf{Chains.} Table~\ref{tab:supp_chain} splits the axis angle of Table~\ref{tab:main} by the structure of the object and of the joint. The error of \ours stays about the same on objects with three or more joints and on trees with three or more levels, while the error of every baseline grows. On joints whose parent is itself a moving link, \ours keeps the lowest error on both sets.

\textbf{Selected fit.} Over the same pools of candidate fits, Table~\ref{tab:supp_select} compares the fit chosen by Eq.~\eqref{ak:eq:select}, which uses only the video, with the expected error of a candidate drawn at random and with the median candidate of each pool. The chosen fit has a lower axis angle than a random candidate on 54 of the 60 objects and on all 8 hands.

\textbf{Articulation from shape alone.} Particulate~\cite{particulate} predicts the parts, tree and joints of a static mesh in one forward pass. We run it with its released checkpoint on the geometry that \ours generates from the first frame, placed in the scene by the same registration, so the two rows of Table~\ref{tab:supp_shape} are two articulation pipelines on the same geometry: one finds the parts, tree and joints from the shape by a learned prior, the other from the video by our fits. Particulate is not given the part count, and a joint it leaves unmatched receives the penalty of Sec.~\ref{sec:supp_protocol}. Particulate predicts joint ranges rather than the observed motion, so we compare the axes. The fits from the video reduce the axis position error by a factor of 2.5 on \harto. On \harth the prior finds one to three joints per hand and leaves 77 of the 93 joints unmatched, while the fits recover the chains.

\end{document}